\documentclass{article}

\usepackage{times}
\usepackage{graphicx} % more modern
\usepackage{subfigure} 

\usepackage{natbib}

\usepackage{algorithm}
\usepackage{algorithmic}
\usepackage{amsmath}
\usepackage{mathrsfs}

\usepackage{pgfplots}
\usepackage{pdfpages}
\usepackage{array,booktabs}% http://ctan.org/pkg/{array,booktabs}

\definecolor{mygreen}{rgb}{0.2, 0.7, 0.2}
\definecolor{myorange}{rgb}{0.9, 0.5, 0.0}

\usepackage{amsbsy}

\newcommand{\T}{\top}
\newcommand{\E}{\mathrm{E}}

\newcommand{\diag}{\mathrm{diag}}
\newcommand{\Tr}{\mathrm{Tr}}
\newcommand{\bigO}{\mathcal{O}}

\newcommand{\norm}{\mathcal{N}}

\newcommand{\avect}{\mathbf{a}}
\newcommand{\bvect}{\mathbf{b}}
\newcommand{\dvect}{\mathbf{d}}

\newcommand{\rvect}{\mathbf{r}}
\newcommand{\fvect}{\mathbf{f}}
\newcommand{\gvect}{\mathbf{g}}

\newcommand{\svect}{\mathbf{s}}
\newcommand{\uvect}{\mathbf{u}}
\newcommand{\vvect}{\mathbf{v}}
\newcommand{\zvect}{\mathbf{z}}
\newcommand{\xvect}{\mathbf{x}}
\newcommand{\yvect}{\mathbf{y}}
\newcommand{\kvect}{\mathbf{k}}
\newcommand{\zerovect}{\mathbf{0}}

\newcommand{\thetavect}{\boldsymbol{\theta}}

\newcommand{\muvect}{\boldsymbol{\mu}}

\usepackage{hyperref}

\usepackage[accepted]{icml2016}

\icmltitlerunning{Preconditioning Kernel Matrices}

\begin{document} 

\twocolumn[
\icmltitle{Preconditioning Kernel Matrices}
% Suggestions for titles: ``Nystr\"om Preconditioned Gaussian Processes'', or simply ``Preconditioned Gaussian Processes'', or ``Gaussian Processes without Cholesky decompositions'', or ``Preconditioning Kernel Matrices''

\icmlauthor{Kurt Cutajar}{kurt.cutajar@eurecom.fr}
\icmladdress{EURECOM,
             Department of Data Science%, 450 Route des Chappes, 06904 Biot, France
}
\icmlauthor{Michael A. Osborne}{mosb@robots.ox.ac.uk}
\icmladdress{University of Oxford,
            Department of Engineering Science%, Parks Road, Oxford, OX1 3PJ, United Kingdom
}
\icmlauthor{John P. Cunningham}{jpc2181@columbia.edu}
\icmladdress{Columbia University,
            Department of Statistics%, 1255 Amsterdam Avenue, New York, NY 10027, USA
}
\icmlauthor{Maurizio Filippone}{maurizio.filippone@eurecom.fr}
\icmladdress{EURECOM,
            Department of Data Science%, 450 Route des Chappes, 06904 Biot, France
}

\icmlkeywords{Gaussian processes, Kernel methods, Conjugate gradient, Stochastic gradient optimization}

\vskip 0.4in
]

\begin{abstract} 
The computational and storage complexity of kernel machines presents the primary barrier to their scaling to large, modern, datasets.  
A common way to tackle the scalability issue is to use the conjugate gradient algorithm, which relieves the constraints on both storage (the kernel matrix need not be stored) and computation (both stochastic gradients and parallelization can be used).  
Even so, conjugate gradient is not without its own issues: the conditioning of kernel matrices is often such that conjugate gradients will have poor convergence in practice.  
Preconditioning is a common approach to alleviating this issue.  
Here we propose preconditioned conjugate gradients for kernel machines, and develop a broad range of preconditioners particularly useful for kernel matrices.  
We describe a scalable approach to both solving kernel machines and learning their hyperparameters.  
We show this approach is exact in the limit of iterations and outperforms state-of-the-art approximations for a given computational budget.
\end{abstract} 

\section{Introduction} \label{sec:introduction}

Kernel machines, in enabling flexible feature space representations of data, comprise a broad and important class of tools throughout machine learning and statistics; prominent examples include support vector machines \cite{Scholkopf01} and Gaussian processes (GPs) \cite{Rasmussen06}.
At the core of most kernel machines is the need to solve linear systems involving the Gram matrix $K = \left\{ k(\xvect_i, \xvect_j \mid \thetavect) \right\}_{i,j=1,...,n}$, where the kernel function $k$, parameterized by $\thetavect$, implicitly specifies the feature space representation of data points $\xvect_i$.   
Because $K$ grows with the number of data points $n$, a fundamental computational bottleneck exists: storing $K$ is $\bigO(n^2)$, and solving a linear system with $K$ is $\bigO(n^3)$.
As the need for large-scale kernel machines grows, much work has been directed towards this scaling issue.

Standard approaches to kernel machines involve a factorization (typically Cholesky) of $K$, which is efficient and exact but maintains the quadratic storage and cubic runtime costs.      
This cost is particularly acute when adapting (or learning) hyperparameters $\thetavect$ of the kernel function, as $K$ must then be factorized afresh for each $\thetavect$.
To alleviate this burden, numerous works have turned to approximate methods \cite{Candela05,Snelson07,Rahimi08} or methods that exploit structure in the kernel~\cite{Gilboa15}.   
Approximate methods can achieve attractive scaling, often through the use of low-rank approximations to $K$, but they can incur a potentially severe loss of accuracy.  
An alternative to factorization is found in the conjugate gradient method (CG), which is used to directly solve linear systems via a sequence of matrix-vector products.  
Any kernel structure can then be exploited to enable fast multiplications, driving similarly attractive runtime improvements, and eliminating the storage burden (neither $K$ nor its factors need be represented in memory). 
Unfortunately, in the absence of special structure that accelerates multiplications, CG performs no better than  $\bigO(n^3)$ in the worst case, and in practice finite numerical precision often results in a degradation of runtime performance compared to factorization approaches.

Throughout optimization, the typical approach to the slow convergence of CG is to apply preconditioners to improve the geometry of the linear system being solved \cite{Golub96}.  
While  preconditioning has been explored in domains such as spatial statistics~\cite{Chen05,Stein12,Ashby96}, the application of preconditioning to kernel matrices in machine learning has received little attention.
Here we design and study preconditioned conjugate gradient methods (PCG) for use in kernel machines, and provide a full exploration of the use of approximations of $K$ as preconditioners. %, first proposed in Davies' PhD thesis \cite{Davies14}. 

Our contributions are as follows.
%% \begin{itemize}
(i) Extending the work in \cite{Davies14}, we apply a broad range of kernel matrix approximations as preconditioners. %, including the Nystr\"om method \cite{Williams00b}, partially- and fully-independent training conditional methods (PITC and FITC) \cite{Snelson05}, randomized partial singular value decomposition (SVD) \cite{Halko11}, inner conjugate gradients with regularization \cite{Srinivasan14}, random Fourier features \cite{Rahimi08,Gredilla10} and structured kernel interpolation (SKI) \cite{Wilson15}.  
    Interestingly, this step allows us to exploit the important developments of \emph{approximate} kernel machines to accelerate the \emph{exact} computation that PCG offers.  
(ii) As a motivating example used throughout the paper, we analyze and provide a general framework to both learn kernel parameters and make predictions in GPs.
(iii) We extend stochastic gradient learning for GPs \cite{FilipponeICML15,Anitescu12} to allow any likelihood that factorizes over the data points by developing an unbiased estimate of the gradient of the approximate log-marginal likelihood.
    We demonstrate this contribution in making the first use of PCG for GP classification.
(iv) We evaluate datasets over a range of problem size and dimensionality. 
    Because PCG is exact in the limit of iterations (unlike approximate techniques), we demonstrate a tradeoff between accuracy and computational effort that improves beyond state-of-the-art approximation and factorization approaches.
%% \end{itemize}

In all, we show that PCG, with a thoughtful choice of preconditioner, is a competitive strategy which is possibly even superior than existing approximation and CG-based techniques for solving general kernel machines\footnote{Code to replicate all results in this paper is available at \\ {\scriptsize \url{http://github.com/mauriziofilippone/preconditioned_GPs}}}.
  
%\noteMF{Other notes:
%Previous theoretical work on the difference between Nystrom and Random Fourier Features in \cite{Yang12}.
%Comparison following error versus time plots as in \cite{Chalupka13}.
%}

\newcommand{\noise}{\lambda I}
\newcommand{\reg}{\delta I}

\section{Motivating example -- Gaussian Processes}\label{sec:GPs}

Gaussian processes (GPs) are the fundamental building block of many probabilistic kernel machines that can be applied in a large variety of modeling scenarios \cite{Rasmussen06}. %, classification \cite{GibbsPhD97,Williams98}, ordinal regression \cite{Chu05}, modeling of count data \cite{Moller98}, and modeling of volatility in time series \cite{Wilson10} to name a few.
%% \noteJC{ this literature review is not necessary.}
Throughout the paper, we will denote by $X = \{\xvect_1, \ldots, \xvect_n\}$ a set of $n$ input vectors and use $\yvect = (y_1, \ldots, y_n)^{\T}$ for the corresponding labels. 
GPs are formally defined as collections of random variables characterized by the property that any finite number of them is jointly Gaussian distributed.
% The random variables in the collection are usually associated with elements in a domain of interest, so that conceptually we can think of GPs as probability distributions over functions.
The specification of a kernel function determines the covariance structure of such random variables
$$
\mathrm{cov}\bigl(f(\xvect), f(\xvect^{\prime})\bigr) = k(\xvect, \xvect^{\prime} \mid \thetavect).
$$
In this work we focus in particular on the popular Radial Basis Function (RBF) kernel % \noteMF{Let's see if we can also add experiments on the Matern or periodic kernels}
\begin{equation} \label{eq:rbf:covariance}
k(\xvect_i, \xvect_j \mid \thetavect) = 
\sigma^2 \exp\left[-\frac{1}{2} \sum_{r=1}^d \frac{ (\xvect_i - \xvect_j)^2_r}{l_r^2} \right],
\end{equation}
where $\thetavect$ represents the collection of the kernel parameters $\sigma^2$ and $l_r^2$. 
% In this definition, we can allow for one length-scale parameter $l_r$ for each feature, which is a suitable modeling assumption for Automatic Relevance Determination (ARD)~\cite{Mackay94}, or for a global length-scale parameter $l$ shared across features.
% The fact that the any finite number of random variables $f(\xvect)$ are jointly Gaussian, makes it possible to obtain a tractable formulation for modeling their probability distribution at a finite number of locations, which in practice include the training input vectors $X$ at training stage, and any test input vectors at prediction stage.
Defining $f_i = f(\xvect_i)$ and $\fvect = (f_1, \ldots, f_n)^{\T}$, and assuming a zero mean GP, we have
$$
\fvect \sim \norm(\fvect \mid \zerovect, K),
$$
where $K$ is the $n \times n$ Gram matrix with elements $K_{ij} = k(\xvect_i, \xvect_j \mid \thetavect)$. 
Note that the kernel above and many popular kernels in machine learning give rise to dense kernel matrices.
%% This kernel in general gives rise to a dense kernel matrix $K$. % The latter approach tends to yield denser kernel matrices.
%GPs form the building block of probabilistic nonparametric models by
% Offering a non-parametric modeling of functions, GPs can be employed to replace the linear component of Generalized Linear Models for added flexibility with the appealing property of maintaining a probabilistic formulation.
Observations are then modeled through a transformation $h$ of a set of GP-distributed latent variables, specifying the model
$$
y_i \sim p \bigl(y_i \mid h(f_i) \bigr), \qquad \fvect \sim \norm(\fvect \mid \zerovect, K).
$$ 
%% In this paper we will focus in particular on the application of  Gaussian likelihood case GPs, and discuss the applicability of the proposed framework to GPs with any factorizing likelihood.

\subsection{The need for preconditioning}

The success of nonparametric models based on kernels hinges on the adaptation of kernel parameters $\thetavect$.
The motivation for preconditioning begins with an inspection of the log-marginal likelihood of GP models with prior $\norm(\fvect \mid \zerovect, K)$.
In Gaussian processes with a Gaussian likelihood $y_i \sim \norm(y_i \mid f_i, \lambda)$, we have analytic forms for
\begin{equation*}
\log[p(\yvect \mid \thetavect, X)] = -\frac{1}{2} \log\left(|K_{\yvect}|\right) - \frac{1}{2} \yvect^{\T} K_{\yvect}^{-1} \yvect + \mathrm{const,}
\end{equation*}
and its derivatives with respect to kernel parameters $\theta_i$,
\begin{equation} \label{eq:gradient:exact}
g_i = 
- \frac{1}{2} \Tr \left( K_{\yvect}^{-1} \frac{\partial K_{\yvect}}{\partial \theta_i} \right) + \frac{1}{2} \yvect^{\T} K_{\yvect}^{-1} \frac{\partial K_{\yvect}}{\partial \theta_i} K_{\yvect}^{-1} \yvect.
\end{equation}
where $K_{\yvect} = K + \lambda I$.
The traditional approach involves factorizing the kernel matrix $K_{\yvect} = L L^{\T}$ using the Cholesky algorithm \cite{Golub96} which costs $\bigO(n^3)$ operations.
After that, all other operations cost $\bigO(n^2)$ except for the trace term in the calculation of $g_i$ which once again requires $\bigO(n^3)$ operations.
Similar computations are required for computing mean and variance predictions for test data~\cite{Rasmussen06}. Note that the solution of a linear system is required for computing the variance at every test point.

This approach is not viable for large $n$ and, consequently, many approaches have been proposed to approximate these computations, thus leading to approximate optimal values for $\thetavect$ and approximate predictions.
Here we investigate the possibility of avoiding approximations altogether, by arguing that for parameter optimization it is sufficient to obtain an unbiased estimate of the gradient $g_i$.
In particular, when such an estimate is available, it is possible to employ stochastic gradient optimization that has strong theoretical guarantees \cite{Robbins51}. 
In the case of GPs, the problematic terms in eq.~\ref{eq:gradient:exact} are the solution of the linear system $K_{\yvect}^{-1} \yvect$ and the trace term.
In this work we make use of a stochastic linear algebra result that allows for an approximation of the trace term,
$$ 
\Tr \left( K_{\yvect}^{-1} \frac{\partial K_{\yvect}}{\partial \theta_i} \right) \approx
\frac{1}{N_{\rvect}} \sum_{i=1}^{N_{\rvect}} {\rvect^{(i)}}^{\T} K_{\yvect}^{-1} \frac{\partial K_{\yvect}}{\partial \theta_i} \rvect^{(i)},
$$ 
where the $N_{\rvect}$ vectors $\rvect^{(i)}$ have components drawn from $\{-1, 1\}$ with probability $1/2$.
Verifying that this is an unbiased estimate of the trace term is straightforward considering that $\E\left( \rvect^{(i)} {\rvect^{(i)}}^{\T} \right) = I$ \cite{GibbsPhD97}.

This result shows that all it takes to calculate stochastic gradients is the ability to efficiently solve linear systems.
Linear systems can be iteratively solved using \emph{conjugate gradient} (CG)~\cite{Golub96}.
The advantage of this formulation is that we can attempt to optimize kernel parameters using stochastic gradient optimization without having to store $K_{\yvect}$ and, given that the most expensive operation is now multiplying the kernel matrix by vectors, only $\bigO(n^2)$ computations are required. 
However, it is well known that the convergence of the CG algorithm depends on the condition number $\kappa(K_{\yvect})$ (ratio of largest to smallest eigenvalues), so the suitability of this approach may also be curtailed if $K_{\yvect}$ is badly conditioned.
%This metric is calculated as a ratio of the largest and smallest eigenvalues, as follows:
%\begin{equation}
%\kappa\left(K\right) = \frac{\lambda_{\mathrm{max}}\left(K\right)}{\lambda_{\mathrm{\min}}\left(K\right)}.
%\end{equation}
To this end, a well-known approach for improving the conditioning of a matrix, which in turn accelerates convergence, is \emph{preconditioning}. 
%% This technique can be incorporated into the CG algorithm by transforming the linear system to be better-conditioned, improving convergence. 
This necessitates the introduction of a preconditioning matrix, $P$, which should be chosen in such a way that $P^{-1}K_{\yvect}$ approximates the identity matrix, $I$.
Intuitively, this can be obtained by setting $P=K_{\yvect}$; however, given that in Preconditioned CG (PCG) we are required to solve linear systems involving $P$, this choice would be no easier than solving the original system.  
Thus we must choose $P$ which approximates $K_{\yvect}$ as closely as possible, but which can also be easily inverted. 
The PCG algorithm is shown in Algorithm~\ref{Alg:Pcg}.

\begin{algorithm}[t]
\caption{The Preconditioned CG Algorithm, adapted from~\cite{Golub96}}
\begin{algorithmic} 
\REQUIRE data \emph{X}, vector $\vvect$, convergence threshold $\epsilon$, initial vector \textbf{x}$_0$, maximum no. of iterations \emph{T}
\STATE $\textbf{r}_0 = \vvect - K_{\yvect}\textbf{x}_0;\;\;\textbf{z}_0 = P^{-1}\textbf{r}_0;\;\;\textbf{p}_0 = \textbf{z}_0$
\FOR{i = 0 : \emph{T}}
\STATE {$\alpha_i = \frac{\textbf{r}_i^T\textbf{z}_i}{\textbf{r}_i^TK_{\yvect}\textbf{z}_i}$}
\STATE $\textbf{x}_{i+1} = \textbf{x}_i + \alpha_i\textbf{p}_i$
\STATE $\textbf{r}_{i+1} = \textbf{r}_i + \alpha_i K_{\yvect}\textbf{p}_i$
\IF {$\|\textbf{r}_{i+1}\| < \epsilon$}
\STATE return \textbf{x} = \textbf{x}$_{i+1}$
\ENDIF
\STATE $\textbf{z}_{i+1}=P^{-1}\textbf{r}_{i+1}$
\STATE $\beta_i = \frac{\textbf{r}_{i+1}^T\textbf{r}_{i+1}}{\textbf{r}_i^T\textbf{r}_i}$
\STATE $\textbf{p}_{i+1} = \textbf{p}_{i+1} + \beta_i\textbf{p}_i$
\ENDFOR
\end{algorithmic}
\label{Alg:Pcg}
\end{algorithm}

\subsection{Non-Gaussian Likelihoods}
When the likelihood $p(y_i \mid f_i)$ is not Gaussian, it is no longer possible to analytically integrate out latent variables.
Instead, techniques such as Gaussian approximations (see, e.g., \cite{Kuss05,Nickisch08}) and methods attempting to characterize the full posterior $p(\fvect, \thetavect \mid \yvect)$ \cite{Murray10b,FilipponeML13} may be required.
Among the various schemes to recover tractability in the case of models with a non-Gaussian likelihood, we choose the Laplace approximation, as we can formulate it in a way that only requires the solution of linear systems.
The GP models we consider assume that the likelihood factorizes across all data points $p(\yvect \mid \fvect) = \prod_{i=1}^n p(y_i \mid f_i)$. % and that the latent variables $\fvect$ are given a zero mean GP prior with kernel $K$.
The use of CG for computing the Laplace approximation has been proposed elsewhere~\cite{Flaxman15}, but we make the first use of preconditioning and stochastic gradient estimation within the Laplace approximation to compute stochastic gradients for non-conjugate models.

Defining $W = - \nabla_{\fvect} \nabla_{\fvect} \log[p(\yvect \mid \fvect)]$ (a diagonal matrix), carrying out the Laplace approximation algorithm, computing its derivatives wrt $\thetavect$, and making predictions, all possess the same computational bottleneck: the solution of linear systems involving the matrix
%% \begin{equation}
$B = I + W^{\frac{1}{2}} K W^{\frac{1}{2}}$~\cite{Rasmussen06}.
%% \end{equation} 
For a given $\thetavect$, each iteration of the Laplace approximation algorithm requires solving one linear system involving $B$ and two matrix-vector multiplications involving $K$; the linear system involving $B$ can be solved using CG or PCG.
The Laplace approximation yields the mode $\hat{\fvect}$ of the posterior over latent variables and offers an approximate log-marginal likelihood in the form:
$$
\log[\hat{p}(\yvect \mid \thetavect, X)] = -\frac{1}{2} \log|B| - \frac{1}{2} \hat{\fvect}^{\T} K^{-1} \hat{\fvect} + \log[p(\yvect \mid \hat{\fvect})]
$$
which poses the same computational challenges as the regression case.
Once again, we therefore seek an alternative way to learn kernel parameters by stochastic gradient optimization based on computing unbiased estimates of the gradient of the approximate log-marginal likelihood.
This is complicated further by the inclusion of an additional ``implicit'' term accounting for the change in the solution given by the Laplace approximation for a change in $\thetavect$.
The full derivation of the gradient is rather lengthy and is deferred to the supplementary material. 
Nonetheless, it is worth noting that the calculation of the exact gradient involves trace terms similar to the regression case that cannot be computed for large $n$, and we unbiasedly estimate these using the stochastic approximation of the trace.

\section{Preconditioning Kernel Matrices}\label{sec:precon}

Here we consider choices for kernel preconditioners, and for the sake of clarity we focus on preconditioners for $K_{\yvect}$.
%In this section, we describe a full investigation of preconditioners for kernel matrices, and discuss how each meets our requirements. 
Unless stated otherwise, we shall consider standard \emph{left} preconditioning, whereby the original problem of solving $K_{\yvect}\zvect = \vvect$ is transformed by applying a preconditioner, $P$, to both sides of this equation. This formulation may thus be expressed as
$
P^{-1}K_{\yvect}\zvect=P^{-1}\vvect.
$

\subsection{Nystr\"om type approximations}

The Nystr\"om method was originally proposed to approximate the eigendecomposition of kernel matrices~\cite{Williams00b}; as a result, it offers a way to obtain a low rank approximation of $K$.
This method selects a subset of $m \ll n$ data (inducing points) collected in the set $U$ which are intended for approximating the spectrum of $K$.
The resulting approximation is 
$
\hat{K} = K_{XU} K_{UU}^{-1} K_{UX}
$
where $K_{UU}$ denotes the evaluation of the kernel function over the inducing points, and $K_{XU}$ denotes the evaluation of the kernel function between the input points and the inducing points.
The resulting preconditioner $P = K_{XU} K_{UU}^{-1} K_{UX} + \lambda I$ can be inverted using the matrix inversion lemma
%% \begin{equation}
$$
P^{-1} \vvect = \lambda^{-1}\left[ I - K_{XU}\left(K_{UU} + K_{UX}K_{XU}\right)^{-1}K_{UX}\right]\vvect,
$$
%% \end{equation}
which has $\bigO(m^3)$ complexity.

\subsubsection{Fully and Partially Independent Training Conditional}
The use of a subset of data for approximating a GP kernel has also been utilized in the fully and partially independent training conditional approaches (FITC and PITC, respectively)  for approximating GP regression~\cite{Candela05}. 
In the former case, the prior covariance of the approximation can be written as follows:
\begin{equation*}
P = K_{XU} K_{UU}^{-1} K_{UX} + \textbf{diag}\left(K - K_{XU} K_{UU}^{-1} K_{UX} \right) + \lambda I.
\end{equation*}
As the name implies, this formulation enforces that the latent variables associated with $U$ are taken to be completely conditionally independent. 
On the other hand, the PITC method extends on this approach by enforcing that although inducing points assigned to a designated block are conditionally dependent on each other, there is no dependence between points placed in different blocks:
\begin{equation*}
P = K_{XU} K_{UU}^{-1} K_{UX} + \textbf{bldiag}\left(K - K_{XU} K_{UU}^{-1} K_{UX} \right) + \lambda I.
\end{equation*}
For the FITC preconditioner, the diagonal resulting from the training conditional can be added to the diagonal noise matrix, and the inversion lemma can be invoked as for the Nystr\"om case. Meanwhile, for the PITC preconditioner, the noise diagonal can be added to the block diagonal matrix, which can then be inverted block-by-block. Once again, matrix inversion can then be carried out as before, where the inverted block diagonal matrix takes the place of $\lambda I$ in the original formulation.

\subsection{Approximate factorization of kernel matrices}

This group of preconditioners relies on approximations to $K$ that factorize as $\hat{K} = \Phi \Phi^{\T}$.
We shall consider different ways of determining $\Phi$ such that $P$ can be inverted at a lower cost than the original kernel matrix $K$. 
Once again, this enables us to employ the matrix inversion lemma, and express the linear system: 
$$
P^{-1} \vvect = % (\hat{K} + \lambda I)^{-1} \vvect = 
(\Phi \Phi^{\T} + \lambda I)^{-1} \vvect
=
%% $$
%% \begin{equation}\label{Eqn:Woodbury}
%% \left(A + UCV\right) = A^{-1}-A^{-1}U\left(C^{-1} + VA^{-1}U\right)VA^{-1}
%% \end{equation}
%% as
%% $$
\lambda^{-1} [I - \Phi (I + \Phi^{\T} \Phi)^{-1} \Phi^{\T}] \vvect.
$$
We now review a few methods to approximate the kernel matrix $K$ in the form $\Phi \Phi^{\T}$.

\subsubsection{Spectral Approximation}

The spectral approach uses random Fourier features for deriving a sparse approximation of a GP \cite{Rahimi08}. 
This approach for GPs was introduced in~\cite{Gredilla10}, and relies on the assumption that stationary kernel functions can be represented as the Fourier transform of non-negative measures. 
As such, the elements of $K$ can be approximated as follows:
$$
\hat{K}_{ij} = \frac{\sigma^2_0}{m}\phi(\xvect_{i})^\T\phi(\xvect_j) = \frac{\sigma^2_0}{m}\sum_{r=1}^{m} \cos\left[2\pi \svect_r^{\T}\left(\xvect_i-\xvect_j\right)\right].
$$
%% $$
%% \phi(x) \!= \!\left[\text{cos(2$\pi$s$_1^\top$\text{x}) sin(2$\pi$s$_1^\top$\text{x}) $\dots$ cos(2$\pi$s$_m^\top$\text{x}) sin(2$\pi$s$_m^\top$}\text{x})\right]^\top
%% $$
In the equation above, the vectors $\svect_r$ denote the \emph{spectral points} (or \emph{frequencies}) which % are sampled from a Gaussian distribution approximating a designated kernel matrix. I
in the case of the RBF kernel % the frequencies 
can be sampled from $\norm\left(\textbf{0}, \frac{1}{4 \pi^2}\Lambda\right)$, 
where $\Lambda = \left[1/l_1^2, \dots, 1/l_n^2\right]$. 
%% This approach lends itself well to preconditioning because while obtaining a good approximation of $K$, the constructed kernel may also be inverted using the matrix inversion lemma leading to an overall computation in $\bigO(m^3)$. 
To the best of our knowledge, this is the first time such an approximation has been considered for the purpose of preconditioning kernel matrices.

\subsubsection{Partial SVD}

Another factorization approach that we consider in this work 
%% This can also be extended to methods involving the explicit factorization of $K$. 
%% In particular, we consider 
is the partial singular value decomposition (SVD) method~\cite{Golub96}.
The SVD method factorizes the original kernel matrix $K$ into $A \Lambda A^{\T}$, where $A$ is a unitary matrix and $\Lambda$ is a diagonal matrix of singular values. 
%% which can itself be directly inverted using the inversion lemma. % However, given that factorizing $K$ completely has cubic complexity, this operation is only feasible for small matrices. 
Here, we shall consider a variation of this technique called \emph{randomized truncated} SVD~\cite{Halko11}, which constructs an approximate low rank SVD factorization of $K$ using random sampling to accelerate computations.
% The decomposition of $K$ using this formulation yields $\Phi = A \hat{\Lambda}^{1/2}$, where $\hat{\Lambda}$ is a diagonal matrix of the exact truncated SVD obtained using randomization.

\subsubsection{Structured Kernel Interpolation (SKI)}

Some recent work on approximating GPs has exploited the fast computation of Kronecker matrix-vector multiplications when inputs are located on a Cartesian grid~\cite{Gilboa15}. Unfortunately, not all datasets meet this requirement, thus limiting the widespread application of Kronecker inference. To this end, SKI~\cite{Wilson15} is an approximation technique which exploits the benefits of the Kronecker product without imposing any requirements on the structure of the training data. In particular, a grid of inducing points, $U$, is constructed, and the covariance between the training data and $U$ is then represented as
$K_{XU} = W K_{UU}.$
In this formulation, $W$ denotes a sparse interpolation matrix for assigning weights to the elements of $K_{UU}$. In this manner, a preconditioner exploiting Kronecker structure can be constructed as
$P = W K_{UU} W^{\T} + \lambda I$.
%% In preconditioned CG, the preconditioner appears in linear systems $\widetilde{P}^{-1} \vvect$ for some vector $\vvect$.
If we consider $V = W / \sqrt{\lambda}$, we can rewrite the (inverse) preconditioner as
%$
%\widetilde{P} = \lambda (V K_{\text{UU}} V^{\T} + I)
%$
%and its inverse can then be written as
$
P^{-1} = \lambda^{-1} (V K_{\text{UU}} V^{\T} + I)^{-1}.
$
Since this can no longer be solved directly, we solve this (inner-loop) linear system using the CG algorithm (all within one iteration of the outer-loop PCG).
% The presence of the identity matrix seems promising in making the system well conditioned; however, the conditioning of this matrix is similar to the original kernel matrix (as it should be if the approximation is accurate).
%As a result, f
For badly conditioned systems, although the complexity of the required matrix-vector multiplications is now much less than $\bigO({n^2})$,  the number of iterations to solve linear systems involving the preconditioner is potentially very large, and could diminish the benefits of preconditioning. %

\subsection{Other approaches}

\subsubsection{Block Jacobi}

An alternative to using a single subset of data involves constructing local GPs over segments of the original data~\cite{Snelson07}. An example of such an approach is the \emph{Block Jacobi} approximation, whereby the preconditioner is constructed by taking a block diagonal of $K$ and discarding all other elements in the kernel matrix. In this manner, covariance is only expressed for points within the same block, as 
$P = \textbf{bldiag}\left(K_{\yvect} + \noise\right).$
The inverse of this block diagonal matrix is computationally cheap (also block diagonal).
However, given that a substantial amount of information contained in the original covariance matrix is ignored, this choice is intrinsically a rather crude approach.

\subsubsection{Regularization}

An appealing feature shared by the aforementioned preconditioners (aside from SKI) is that their structure enables us to directly solve $P^{-1} \vvect$. 
An alternative technique for constructing a preconditioner involves adding a positive regularization parameter, $\reg$, to the original kernel matrix, such that $P=K_{\yvect}+\reg$~\cite{Srinivasan14}. 
This follows from the fact that adding noise to the diagonal of $K_{\yvect}$ makes it better-conditioned, and the condition number is expected to decrease further as $\delta$ increases.
Nonetheless, for the purpose of preconditioning, this parameter should be tuned in such a way that $P$ remains a sensible approximation of $K_{\yvect}$. 
As opposed to the previous preconditioners, this is an instance of \emph{right} preconditioning, which has the following general form
$K_{\yvect} P^{-1}(P\textbf{x})=\textbf{v}.$

Given that it is no longer possible to evaluate $P^{-1} \vvect$ analytically, this linear system is solved yet again using CG, such that a linear system of equations is solved at every outer iteration of the PCG algorithm. 
Due to the potential loss of accuracy incurred while solving the inner linear systems, a variation of the standard PCG algorithm, referred to as \emph{flexible} PCG~\cite{Notay00}, is used instead. 
Using this approach, a re-orthogonalization step is introduced such that the search directions remain orthogonal even when the inner system is not solved to high precision.

\section{Comparison of Preconditioners}

\begin{figure*} 
\begin{center}
\renewcommand{\arraystretch}{0.5}
\begin{tabular}{c c c m{1cm}}
{\bf {\footnotesize Concrete dataset}} & \hspace{0.5cm} {\bf {\footnotesize Power plant dataset}} & \hspace{0.5cm} {\bf {\footnotesize Protein dataset}} & \hspace{0.5cm} \\

\scalebox{1}{
\begin{tabular}{c}
\includegraphics[page=1,width=.15\textwidth]{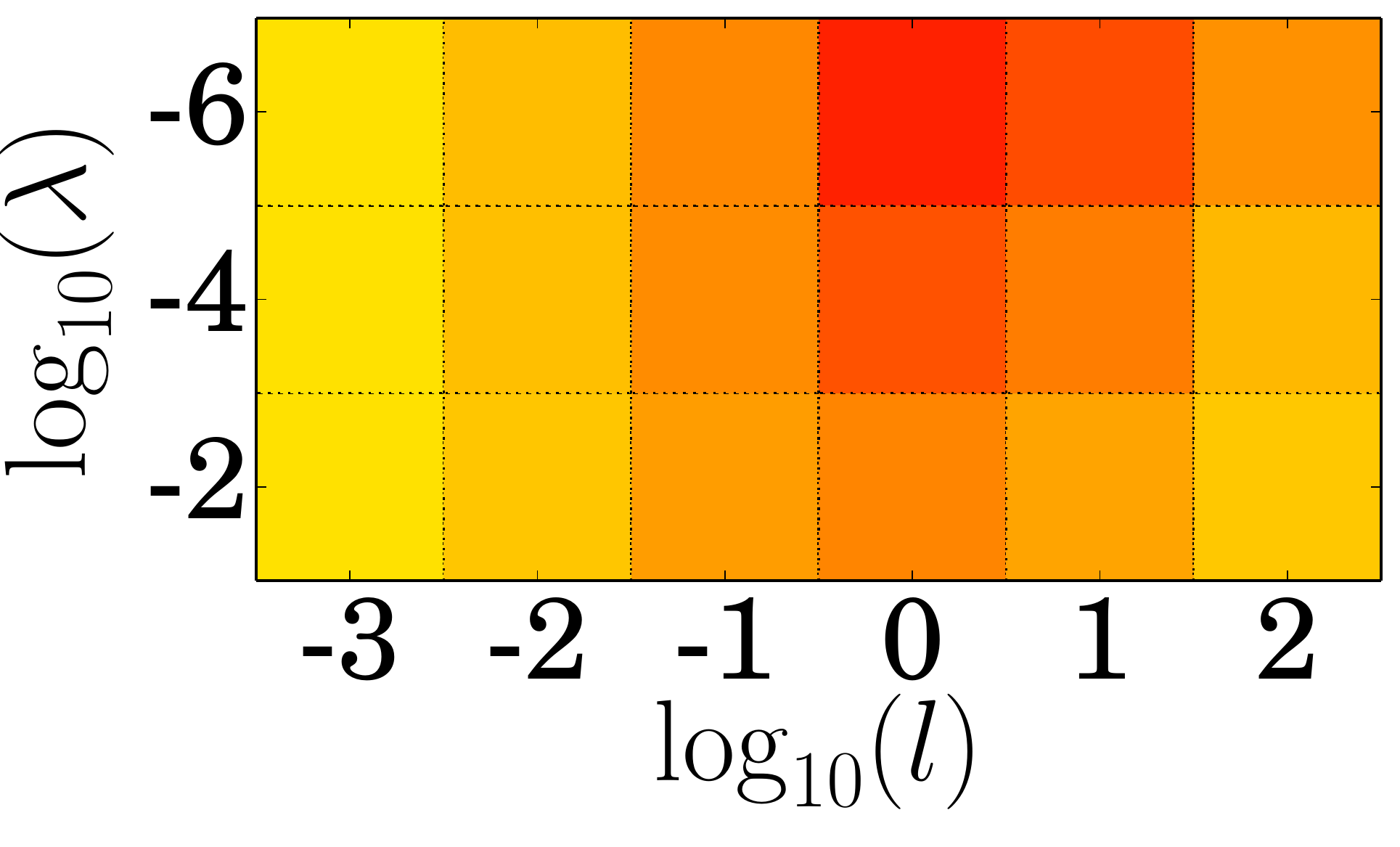}
\end{tabular}} & \hspace{0.5cm}
\scalebox{1}{
\begin{tabular}{c}
\includegraphics[page=1,width=.15\textwidth]{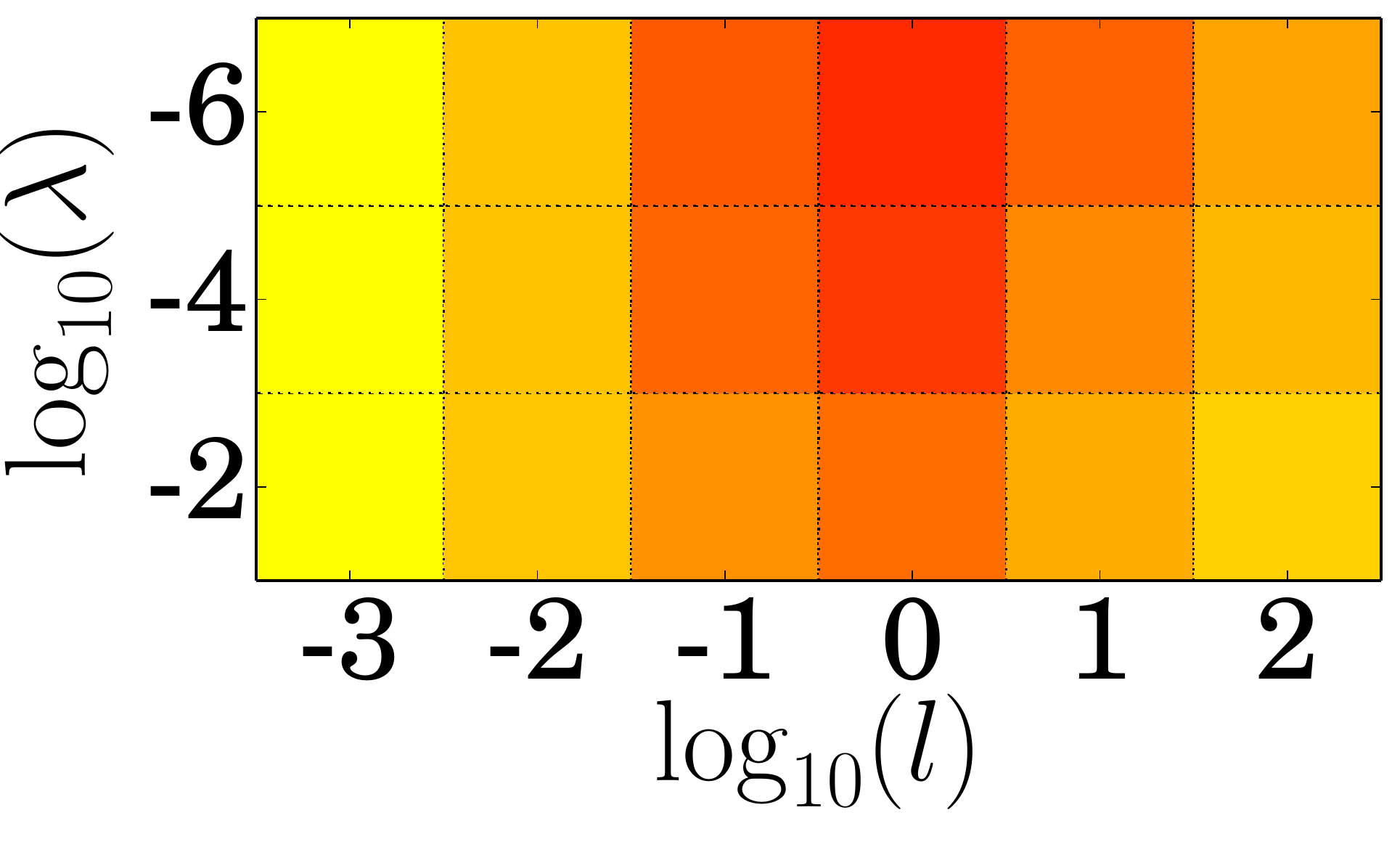}
\end{tabular}} & \hspace{0.5cm}
\scalebox{1}{
\begin{tabular}{c}
\includegraphics[page=1,width=.15\textwidth]{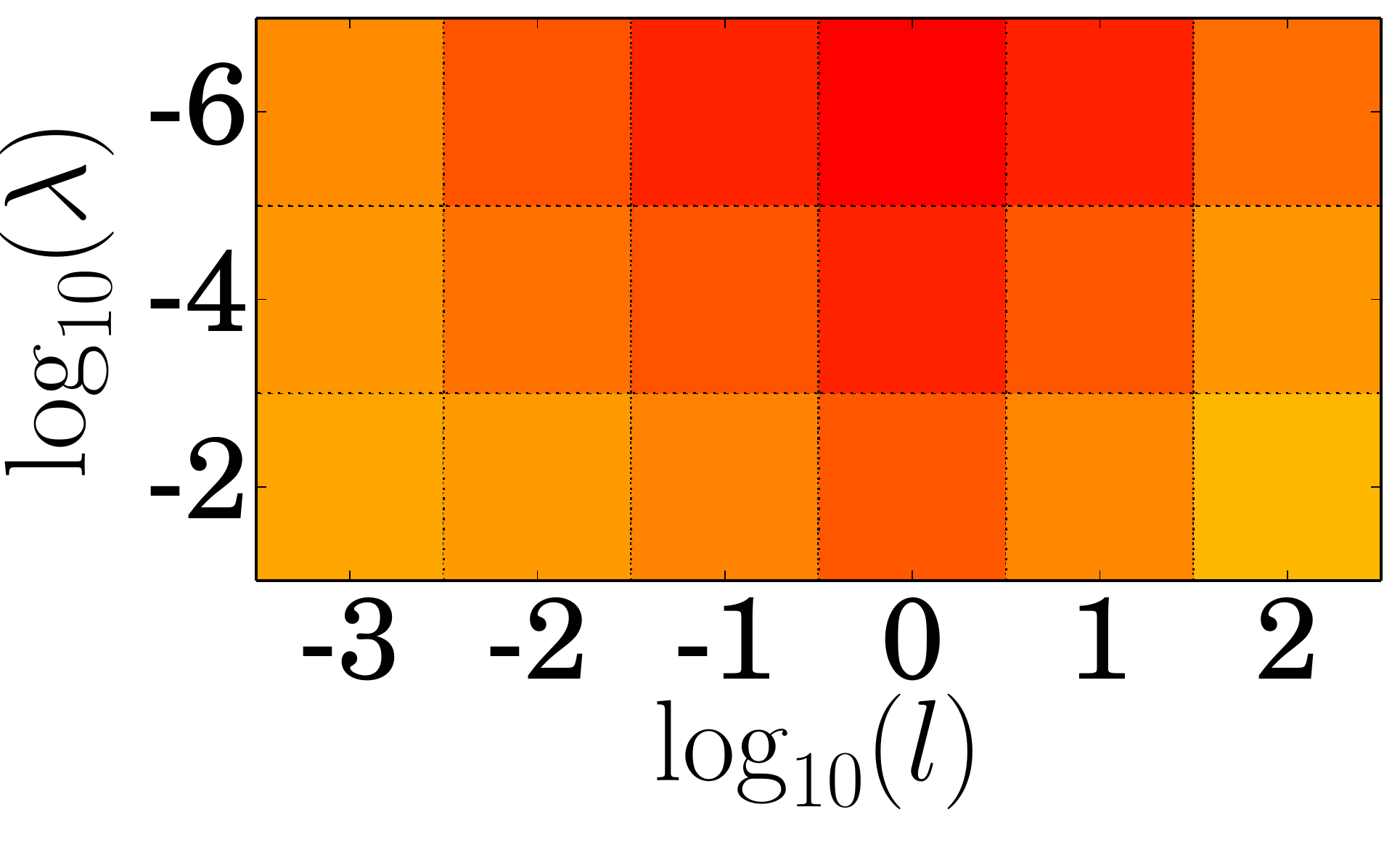}
\end{tabular}} & \hspace{0.5cm}
\begin{tabular}{c}
$\text{log}_{10}(n_\text{it})$ \\
\begin{tikzpicture}
\begin{axis}[
hide axis,
scale only axis,
height=0pt,
width=0pt,
colorbar,
point meta min=0,
point meta max=5,
colormap={mygreen}{rgb255=(255,255,0);rgb255=(255,165,0);rgb255=(255,0,0)},
colorbar style={        
ytick={0,5},
 yticklabel={},
    height=1.2cm,
    y tick scale label style={xshift=0.5cm},
    width=0.3cm
}]
\addplot [draw=none] coordinates {(0,0)};
\end{axis}
\end{tikzpicture} 
\end{tabular}\vspace{0.2cm} \\
\scalebox{0.8}{\begin{tabular}{@{}c@{\hspace{.00001cm}\vspace{0.00005cm}}c@{}}
       \includegraphics[page=2,width=.15\textwidth]{Plots/concreteComp.pdf} & 
       \includegraphics[page=3,width=.15\textwidth]{Plots/concreteComp.pdf} \\
       \includegraphics[page=4,width=.15\textwidth]{Plots/concreteComp.pdf} &
       \includegraphics[page=5,width=.15\textwidth]{Plots/concreteComp.pdf} \\
       \includegraphics[page=6,width=.15\textwidth]{Plots/concreteComp.pdf} &
       \includegraphics[page=7,width=.15\textwidth]{Plots/concreteComp.pdf} \\
       \includegraphics[page=8,width=.15\textwidth]{Plots/concreteComp.pdf} &
       \includegraphics[page=9,width=.15\textwidth]{Plots/concreteComp.pdf} 
\end{tabular}} & \hspace{0.5cm}
\scalebox{0.8}{\begin{tabular}{@{}c@{\hspace{.005cm}\vspace{0.00005cm}}c@{}}
       \includegraphics[page=2,width=.15\textwidth]{Plots/powerComp.pdf} & 
       \includegraphics[page=3,width=.15\textwidth]{Plots/powerComp.pdf} \\
       \includegraphics[page=4,width=.15\textwidth]{Plots/powerComp.pdf} &
       \includegraphics[page=5,width=.15\textwidth]{Plots/powerComp.pdf} \\
       \includegraphics[page=6,width=.15\textwidth]{Plots/powerComp.pdf} &
       \includegraphics[page=7,width=.15\textwidth]{Plots/powerComp.pdf} \\
       \includegraphics[page=8,width=.15\textwidth]{Plots/powerComp.pdf} &
       \includegraphics[page=9,width=.15\textwidth]{Plots/powerComp.pdf} 
\end{tabular}}& \hspace{0.5cm}
\scalebox{0.8}{\begin{tabular}{@{}c@{\hspace{.005cm}\vspace{0.00005cm}}c@{}}
       \includegraphics[page=2,width=.15\textwidth]{Plots/proteinComp.pdf} & 
       \includegraphics[page=3,width=.15\textwidth]{Plots/proteinComp.pdf} \\
       \includegraphics[page=4,width=.15\textwidth]{Plots/proteinComp.pdf} &
       \includegraphics[page=5,width=.15\textwidth]{Plots/proteinComp.pdf} \\
       \includegraphics[page=6,width=.15\textwidth]{Plots/proteinComp.pdf} &
       \includegraphics[page=7,width=.15\textwidth]{Plots/proteinComp.pdf} \vspace{2cm}
\end{tabular}}  & \hspace{0.5cm}
\begin{tabular}{m{1cm}}
\centering {\small Loss}
\\ \begin{tikzpicture}%[xshift=2cm]
\begin{axis}[
hide axis,
scale only axis,
height=0pt,
width=0pt,
colorbar,
point meta min=-2,
point meta max=2,
colormap={mygreen}{rgb255=(0,0,255);color=(white);rgb255=(255,0,0)},
colorbar style={
 yticklabel={},
    height=4.75cm,
    width=0.3cm
}]
\addplot [draw=none] coordinates {(0,0)};
\end{axis}
\end{tikzpicture}
\\\centering {\small Gain}
\end{tabular}
\end{tabular}
\end{center}
\caption{Comparison of preconditioners for different settings of kernel parameters.  
The lengthscale $l$ and the noise variance $\lambda$ are shown on the x and y axes respectively.
The top figure indicates the number of iterations required to solve the corresponding linear system using CG, whilst the bottom part of the figure shows the rate of improvement (negative - blue) or degradation (positive - red) achieved by using PCG to solve the same linear system.
Parameters and results are reported in $\log_{10}$.
Symbols % within cells % $+$, $-$, and $\circ$ % have been 
added to facilitate reading in B/W print.
}
\label{fig:comparison:preconditioners}
\end{figure*}

In this section, we provide an empirical exploration of these preconditioners in a practical setting. 
We begin by considering three datasets for regression from the UCI repository \cite{Asuncion07}, namely the Concrete dataset ($n=1030, d=8$), the Power Plant dataset ($n=9568, d=4$), and the Protein dataset ($n=45730, d=9$).
In particular, we evaluate the convergence in solving $K_{\yvect}\zvect=\yvect$ using iterative methods, where $\yvect$ denotes the labels of the designated  dataset, and $K_{\yvect}$ is constructed using different configurations of kernel parameters.

With this experiment, we aim to assess the quality of different preconditioners based on how many matrix-vector products they require, which, for most approaches, corresponds to the number of iterations taken by PCG to converge.
The convergence threshold is set to $\epsilon^2 = n \cdot 10^{-10}$ so as to roughly accept an average error of $10^{-5}$ on each element of the solution. 

For every variation, we set the parameters of the preconditioners so as to have a complexity lower than the $\bigO(n^2)$ cost associated with matrix-vector products; by doing so, we can assume that the latter computations are the dominant cost for large $n$. 
In particular, for Nystr\"om-type methods, we set $m = \sqrt{n}$ inducing points, so that when we invert the preconditioner using the matrix inversion lemma, the cost is in $\bigO(m^3) = \bigO(n^{3/2})$.
Similarly, for the Spectral preconditioner, we set $m = \sqrt{n}$ random features.
For the SKI preconditioner, we take an equal number of elements on the grid for each dimension; under this assumption, Kronecker products have $\bigO(d n^{\frac{d+1}{d}})$ cost \cite{Gilboa15}, and we set the size of the grid so that the complexity of applying the preconditioner matches $\bigO(n^{3/2})$, so as to be consistent with the other preconditioners.
For the Regularized approach, each iteration needed to apply the preconditioner requires one matrix-vector product, and we add this to the overall count of such computations. 
For this preconditioner, we add a diagonal offset $\delta$ to the original matrix, equivalent to two orders of magnitude greater than the noise of the process.
In general, although the complexity of PCG is indeed no different from that of CG, we emphasize that experiencing a 2-fold or 5-fold (in some cases even an order of magnitude) improvement can be very substantial when plain CG takes very long to converge or when the dataset is large.

We focus on an isotropic RBF variant of the kernel in eq.~\ref{fig:comparison:preconditioners}, fixing the marginal variance $\sigma^2$ to one.
We vary the length-scale parameter $l$ and the noise variance $\lambda$ in $\log_{10}$ scale.
The top part of fig.~\ref{fig:comparison:preconditioners} shows the number of iterations that the standard CG algorithm takes, where we have capped the number of iterations to 100,000.

The bottom part of the figure reports the improvement offered by various preconditioners measured as
$$
\log_{10}{\left(\frac{\#\ \mathrm{PCG\ iterations}}{\#\ \mathrm{CG\ iterations}} \right)}.
$$
It is worth noting that when both CG and PCG fail to converge within the upper bound, the improvement will be marked as 0, i.e. neither a gain or a loss within the given bound.
The results plotted in fig.~\ref{fig:comparison:preconditioners} indicate that the low-rank preconditioners (PITC, FITC and Nystr\"{o}m) achieve significant reductions in the number of iterations for each dataset, and all approaches work best when the lengthscale is longer, characterising smoother processes. 
In contrast, preconditioning seems to be less effective when the lengthscale is shorter, corresponding to a kernel matrix that is more sparse. 
However, for cases yielding positive results, the improvement is often in the range of an order of magnitude, which can be  substantial when a large number of iterations is required by the CG algorithm. 
%% Although all preconditioners perform similarly across different regions of the chosen grid, the Nystr\"{o}m method frequently perform better than the rest.

The results also confirm that, as alluded to in the previous section, Block Jacobi preconditioning is generally a poor preconditioner, particularly when the corresponding kernel matrix is dense. 
The only minor improvements were observed when CG itself converges quickly, in which case preconditioning serves very little purpose either way. 

The regularization approach with flexible conjugate gradient does not appear to be effective in any case either, particularly due to the substantial amount of iterations required for solving an inner system at every iteration of the PCG algorithm. 
This implies that introducing additional small jitter to the diagonal does not necessarily make the system much easier to solve, whilst adding an overly large offset would negatively impact convergence of the outer algorithm. 
One could assume that tuning the value of this parameter could result in slightly better results; however, preliminary experiments in this regard yielded only minor improvements.

The results for SKI preconditioning are similarly discouraging at face value. 
%% Given that inner matrix-vector products can exploit Kronecker structure, we permitted the upper limit of 15,000 outer iterations. 
%% However, it transpired that w
When the matrix $K_{\yvect}$ is very badly conditioned, an excessive number of inner iterations are required for every iteration of outer PCG. 
This greatly increases the duration of solving such systems, and as a result, this method was not included in the comparison for the Protein dataset, where it was evident that preconditioning the matrix in this manner would not yield satisfactory improvements. 
Notwithstanding that these experiments depict a negative view of SKI preconditioning, it must be said that we assumed a fairly simplistic interpolation procedure in our experiments, where each data point was mapped to nearest grid location. 
The size of the constructed grid is also hindered considerably by the constraint imposed by our upper bound on complexity. 
Conversely, more sophisticated interpolation strategies or even grid formulation procedures could possibly speed up the convergence of CG for the inner systems. 
In line with this thought, however, one could argue that the preconditioner would no longer be straightforward to construct, which goes against our innate preference towards easily derived preconditioners.

\section{Impact of preconditioning on GP learning}

\begin{figure*}[ht]
% \vskip 0.2in
\begin{center}
\begin{tabular}{cccc}
%% \multicolumn{4}{c}{{\bf Concrete dataset - regression - $n \approx 1K$}} \\
%% \multicolumn{2}{c}{{\bf Isotropic}} &
%% \multicolumn{2}{c}{{\bf ARD}} \\
%% \includegraphics[width=\columnwidth/2]{Plots/PLOT_concrete_RBF_RMSE.pdf} \hspace{-0.5cm} & \includegraphics[width=\columnwidth/2]{Plots/PLOT_concrete_RBF_NEG_LLIK.pdf} &
%% \includegraphics[width=\columnwidth/2]{Plots/PLOT_concrete_ARD_RMSE.pdf} \hspace{-0.5cm} & \includegraphics[width=\columnwidth/2]{Plots/PLOT_concrete_ARD_NEG_LLIK.pdf} \\
%% \multicolumn{4}{c}{{\bf Power Plant dataset - regression - $n \approx 10K$}}\\
%% \multicolumn{2}{c}{{\bf Isotropic}} &
%% \multicolumn{2}{c}{{\bf ARD}} \\
\includegraphics[width=\columnwidth/2]{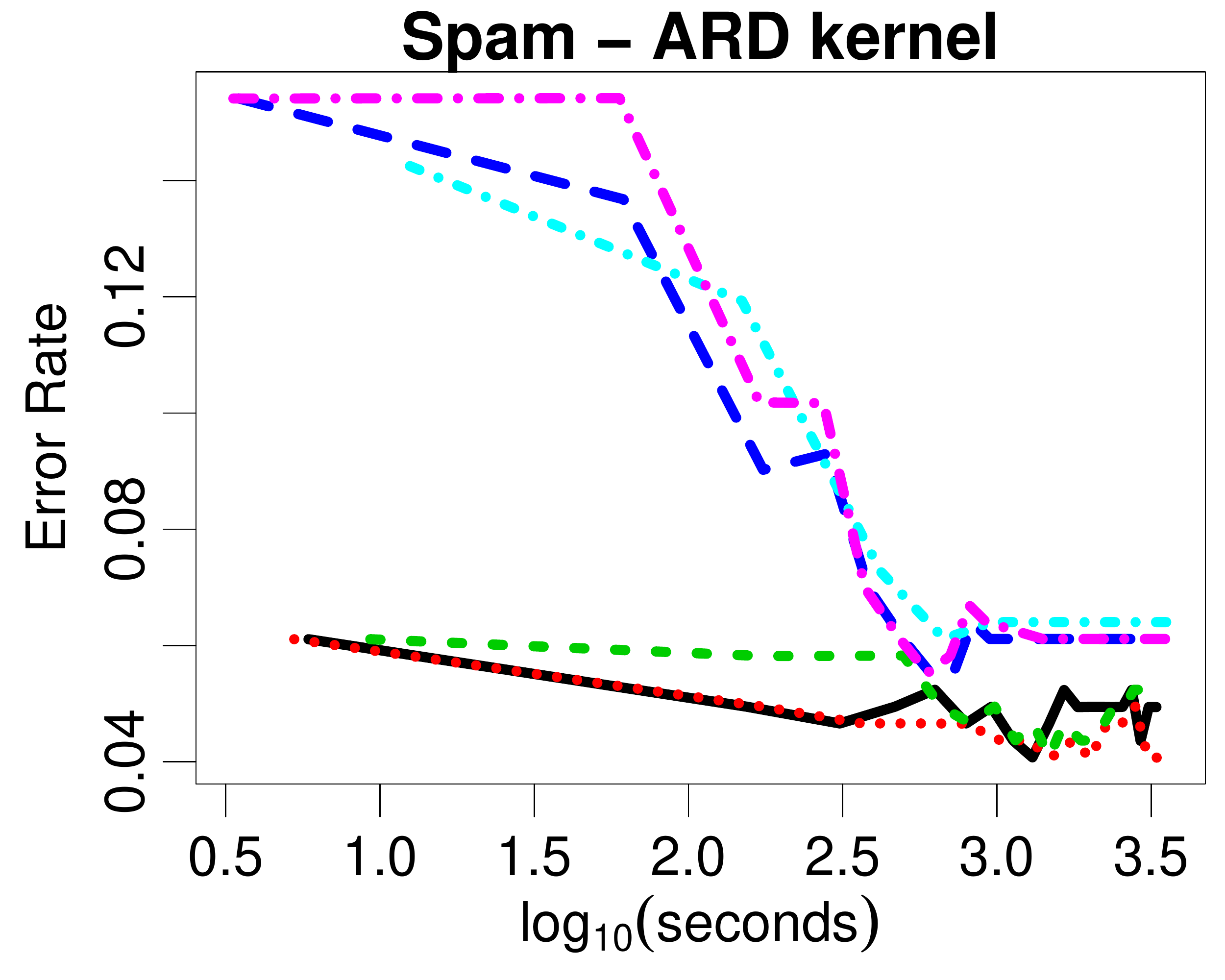} \hspace{-0.5cm} & \includegraphics[width=\columnwidth/2]{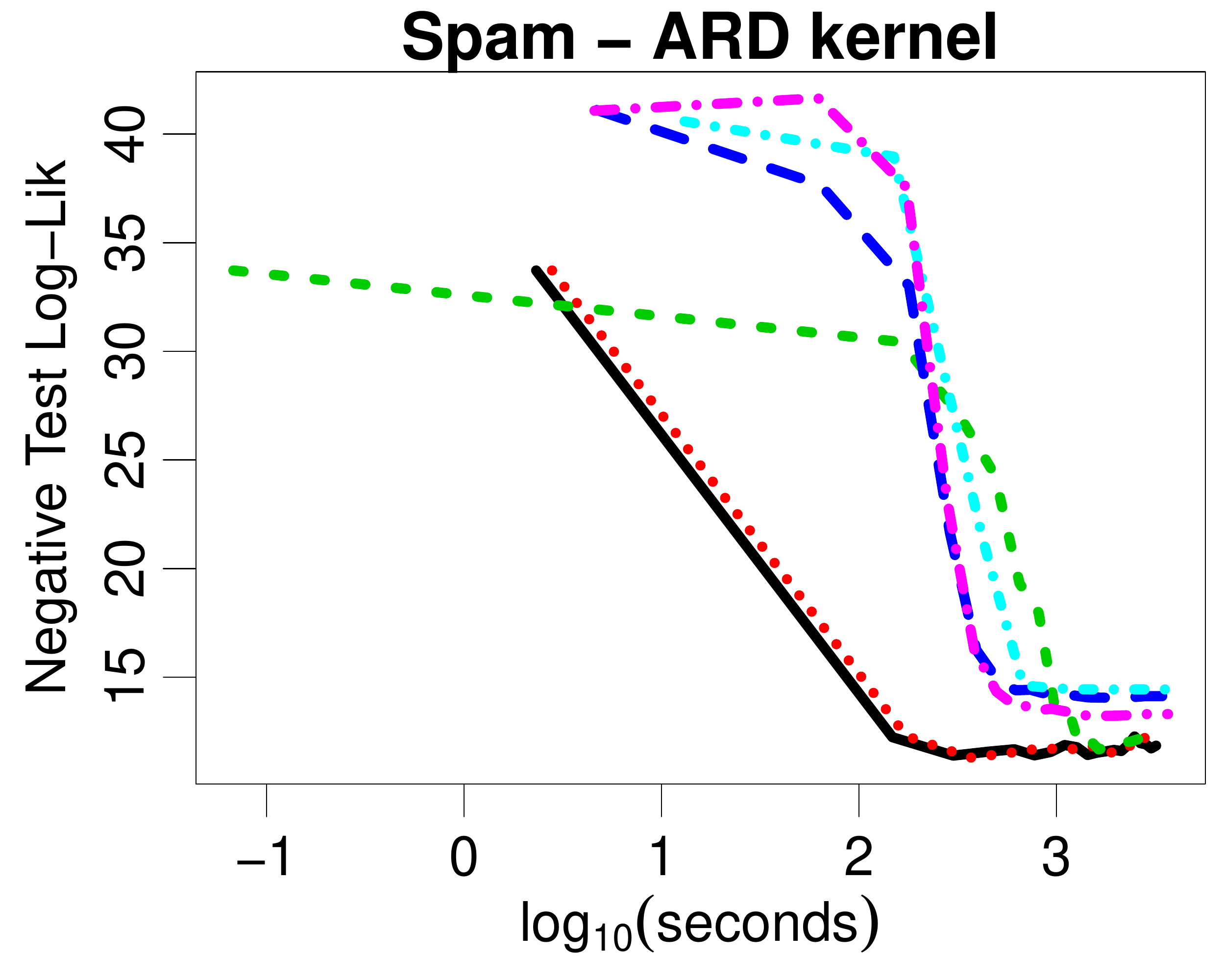} &
\includegraphics[width=\columnwidth/2]{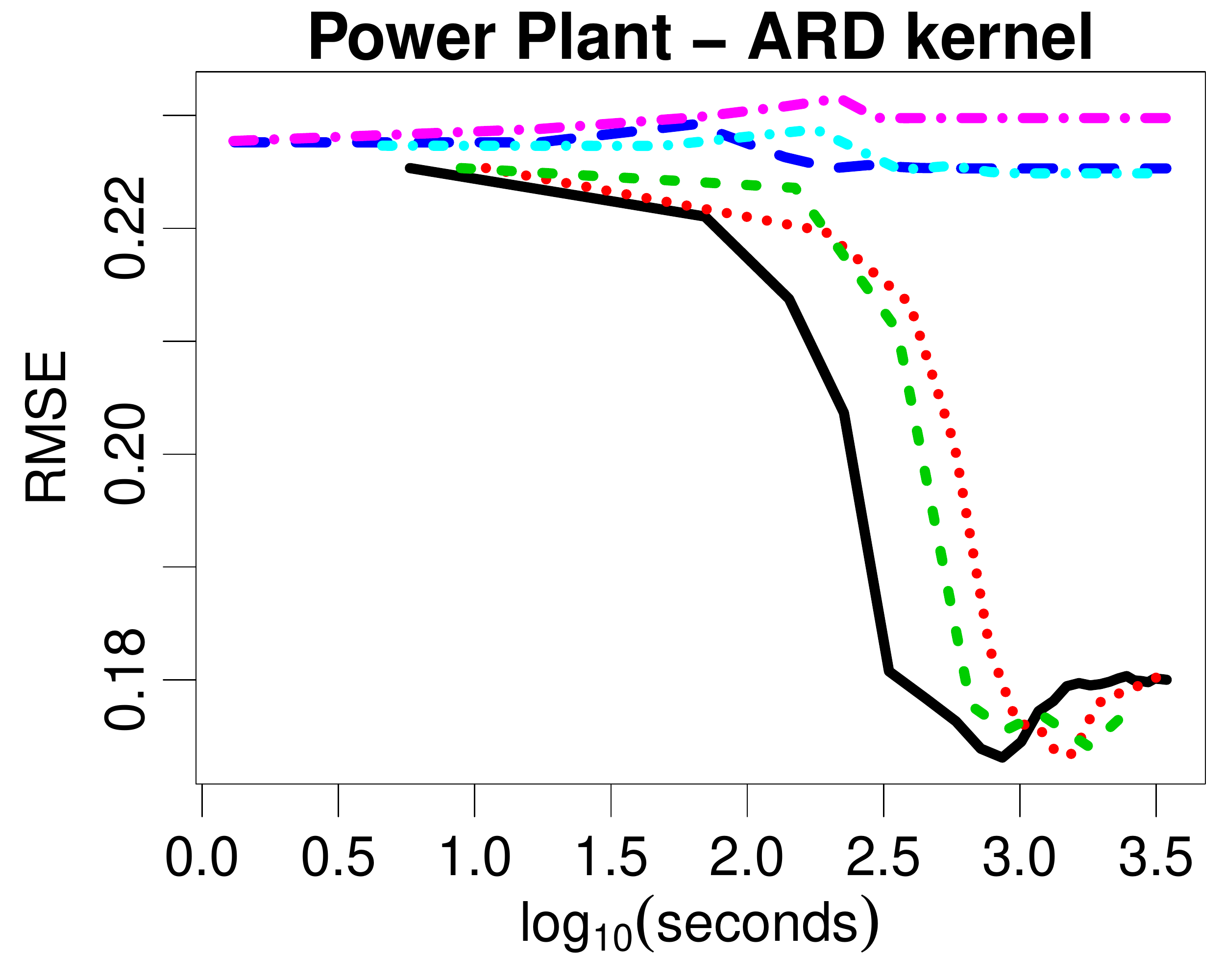} \hspace{-0.5cm} & \includegraphics[width=\columnwidth/2]{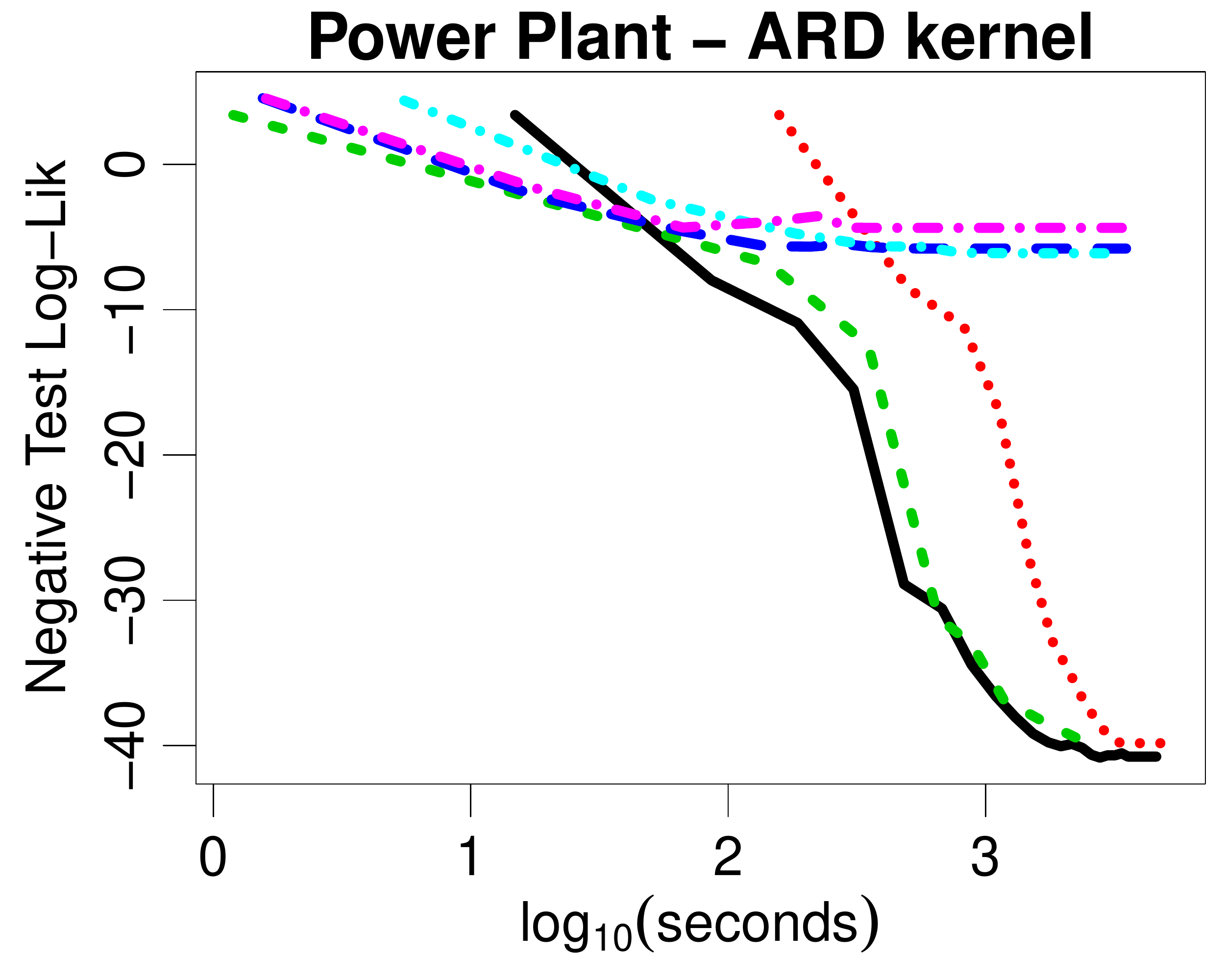} \\
%% \multicolumn{4}{c}{{\bf Protein dataset - regression - $n \approx 45K$}}\\
%% \multicolumn{2}{c}{{\bf Isotropic}} &
%% \multicolumn{2}{c}{{\bf ARD}} \\
\includegraphics[width=\columnwidth/2]{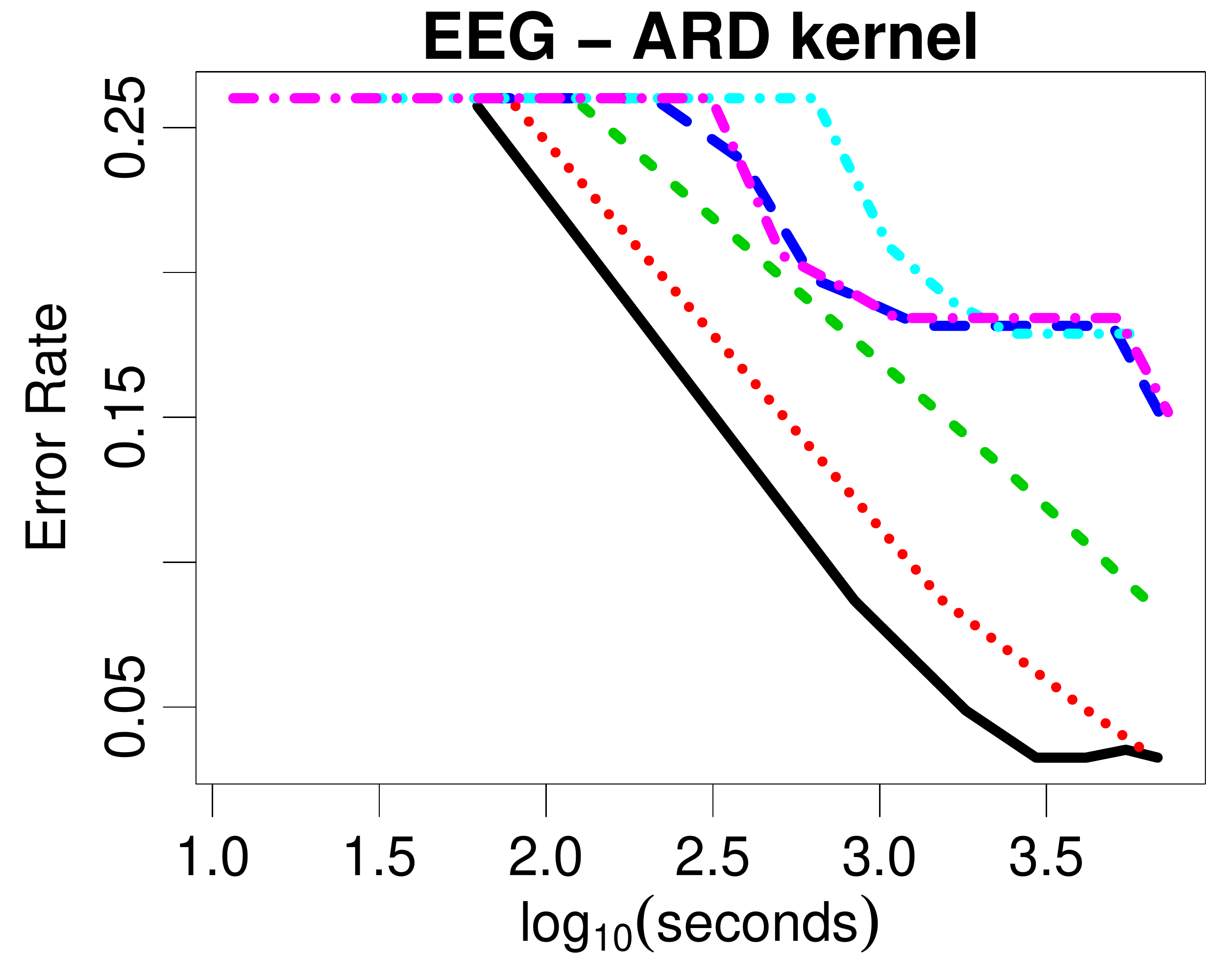} \hspace{-0.5cm} & \includegraphics[width=\columnwidth/2]{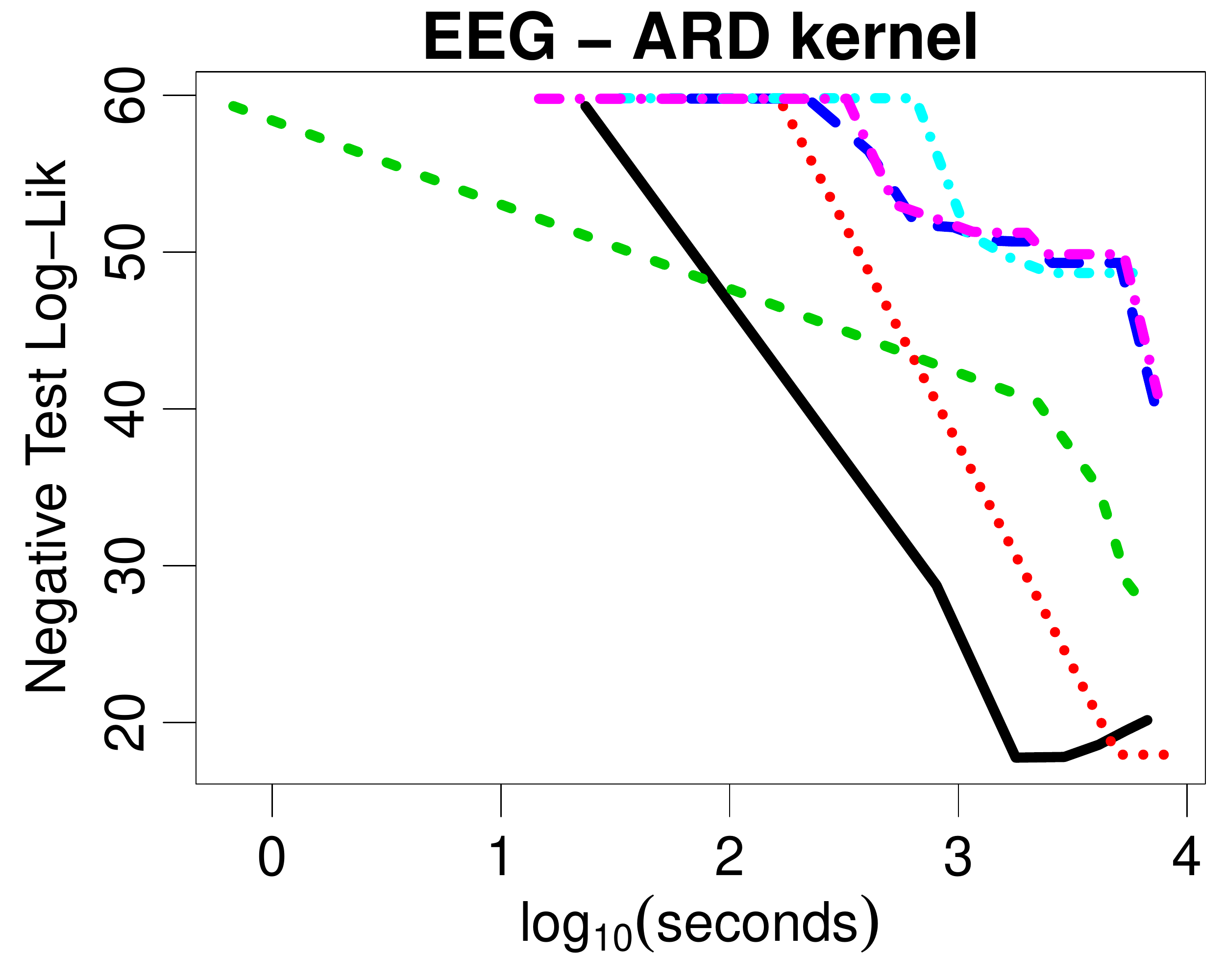} &
\includegraphics[width=\columnwidth/2]{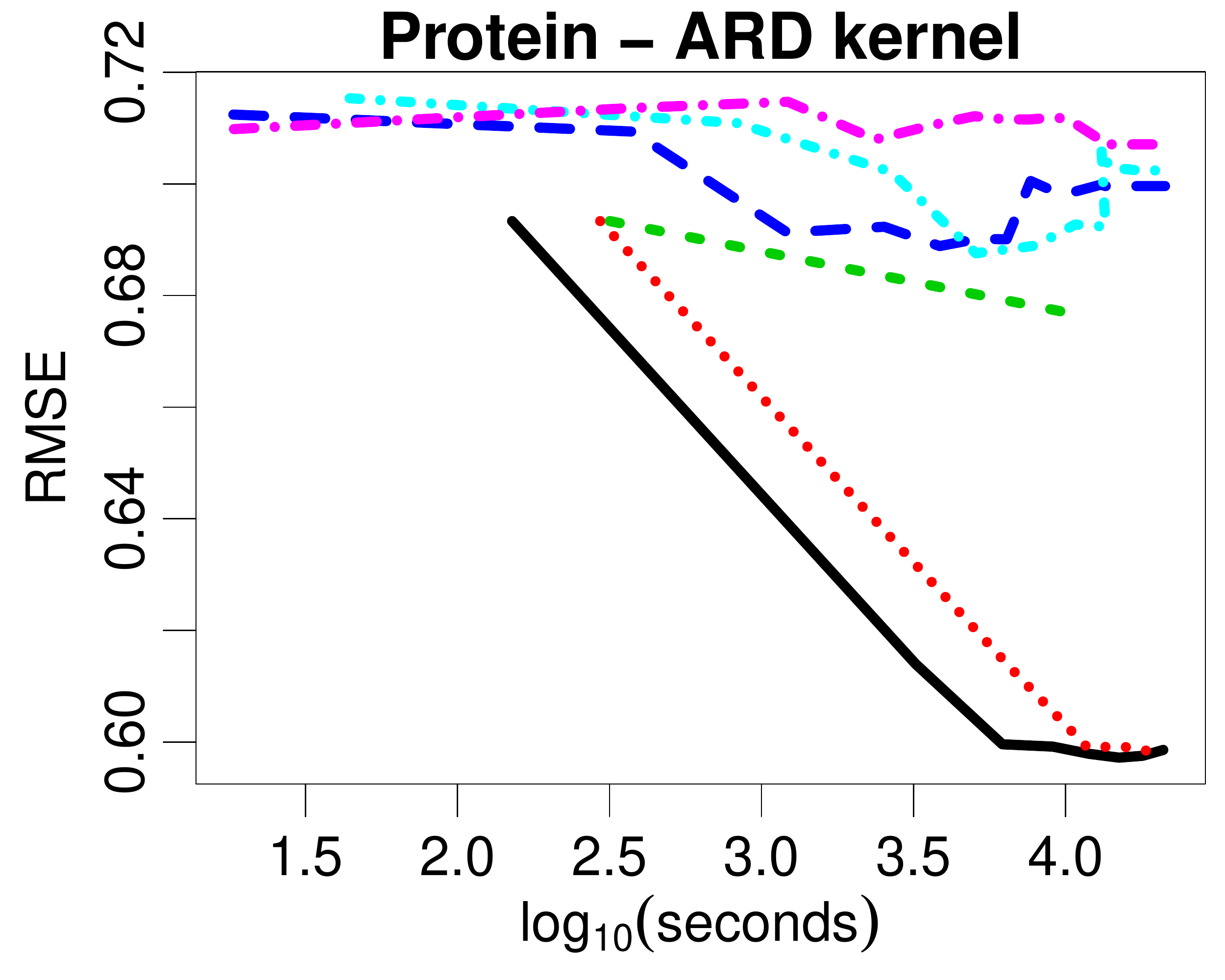} \hspace{-0.5cm} & \includegraphics[width=\columnwidth/2]{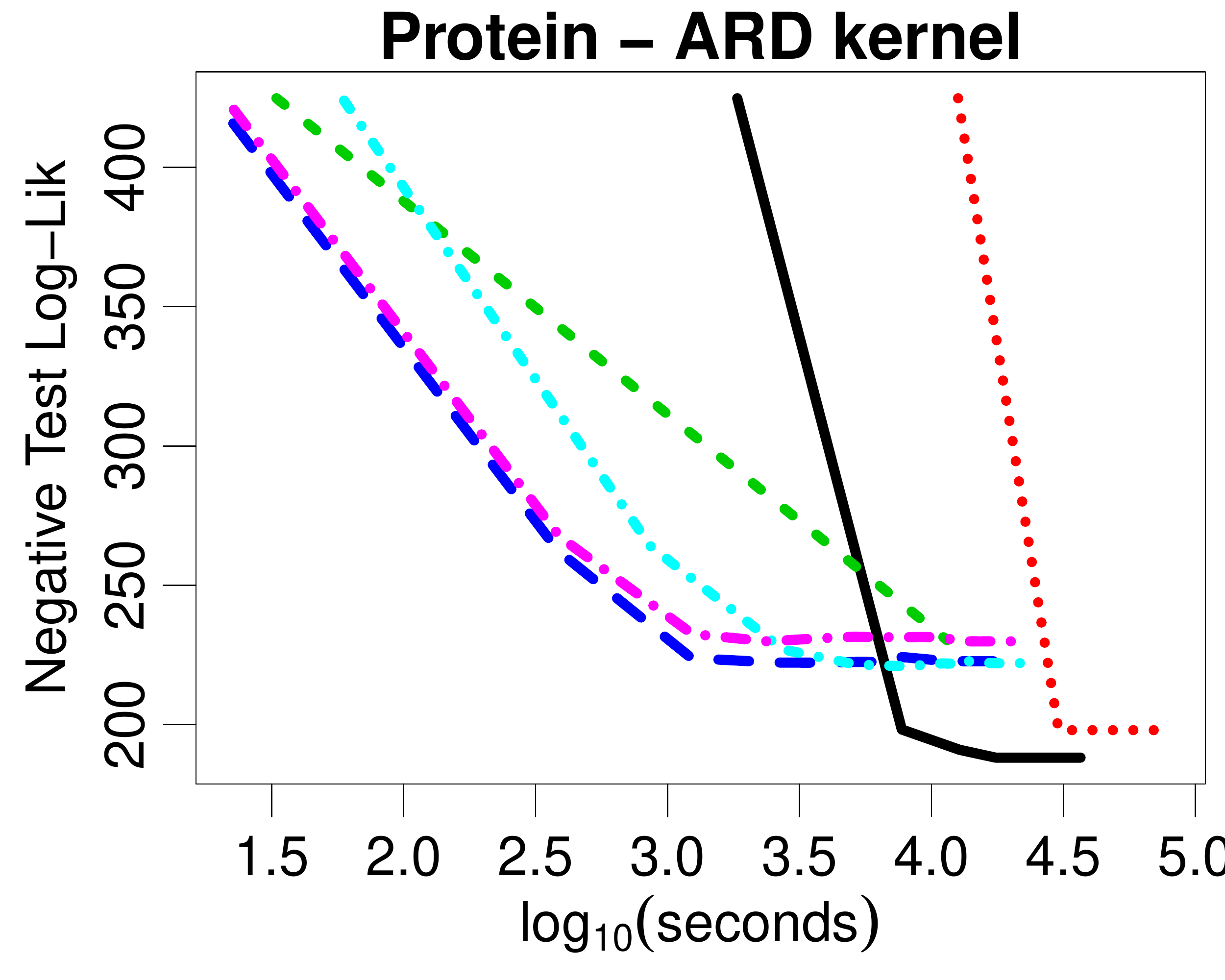} \\
\multicolumn{4}{c}{\includegraphics[width=\columnwidth]{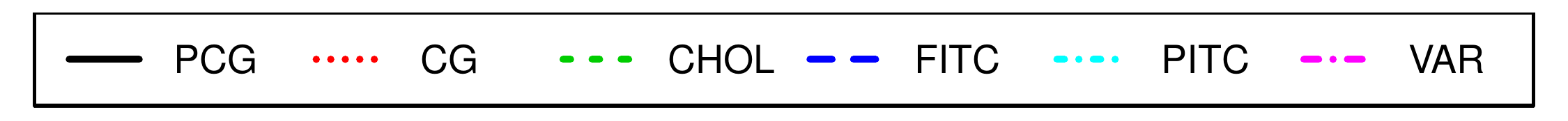}} 
\end{tabular}
\caption{RMSE and negative log of the likelihood on $\sqrt{n}$ held out test data over time. GP models employ the ARD kernel in eq.~\ref{eq:rbf:covariance}. GP classification: Spam dataset ($n=4601, d=57$) and EEG dataset ($n=14979, d=14$). GP regression: Power Plant dataset ($n=9568, d=4$) and Protein dataset ($n=45730, d=9$). Curves are averaged over multiple repetitions.}
\label{fig:error_vs_time}
\end{center}
\vskip -0.2in
\end{figure*} 

One of the primary objectives of this work is to reformulate GP regression and classification in such a way that preconditioning can be effectively exploited. 
In section~\ref{sec:GPs}, we demonstrated how preconditioning can indeed be applied to GP regression problems, and also proposed a novel way of rewriting GP classification in terms of solving linear systems (where preconditioning can thus be employed). 
We can now evaluate how the proposed preconditioned GP techniques  compare to other state of the art methods.

To this end, in this section, we empirically report on the generalization ability of GPs as a function of the time taken to optimize parameters $\thetavect$ and compute predictions.
In particular, for each of the methods featured in our comparison, we iteratively run the optimization of kernel parameters for a few iterations and predict on unseen data, and assess how prediction accuracy varies over time for different methods. 

The analysis provided in this report is inspired by \cite{Chalupka13}, although we do not propose an \emph{approximate} method to learn GP kernel parameters. Instead, we put forward a means of accelerating the optimization of kernel parameters \emph{without any approximation}\footnote{%
    The one proviso to this statement is that, for non-Gaussian likelihood, stochastic gradients target the approximate log-marginal likelihood obtained by the Laplace approximation.   
}.
Given the predictive mean and variance for the $N_{\mathrm{test}}$ test points, say $m_{*i}$ and  $s^2_{*i}$, we report two error measures, namely the Root Mean Square Error,
$
\mathrm{RMSE} = \sqrt{\frac{1}{N_{\mathrm{test}}} \sum_{i = 1}^{N_{\mathrm{test}}} (m_{*i} - y_{*i})^2 },
$
along with the negative log-likelihood on the test data,
$
- \sum_{i = 1}^{N_{\mathrm{test}}} \log[p(y_{*i} \mid m_{*i}, s^2_{*i})],
$
where $y_{*i}$ denotes the label of the $i$th of $N_{\mathrm{test}}$ data points.
For classification, instead of the RMSE we report the error rate of the classifier.

We can make use of stochastic gradients for GP models to optimize kernel parameters using off-the-shelf stochastic gradient optimization algorithms.
In order to reduce the number of parameters to tune, we employ ADAGRAD~\cite{Duchi11} -- an optimization algorithm having a single step-size parameter. 
For the purpose of this experiment, we do not attempt to optimize this parameter, since this would require additional computations.
Nonetheless, our experience with training GP models indicates that the choice of this parameter is not critical: we set the step-size to one.

Fig.~\ref{fig:error_vs_time} shows the two error measures over time for a selection of approaches.
In the figure, PCG and CG refer to stochastic gradient optimization of kernel parameters using ADAGRAD, where linear systems are solved with PCG and CG, respectively.
In view of the results obtained in our comparison of preconditioners, we decide to proceed with the Nystr\"om preconditioning method.
Furthermore, we construct the preconditioner with $m = 4 \sqrt{n}$ points randomly selected from the input data at each iteration, such that the overall complexity of the PCG method matches plain CG.
For these methods, stochastic estimates of trace terms are carried out using $N_{\rvect} = 4$ random vectors.
The baseline CHOL method refers to the optimization of kernel parameters using the L-BFGS algorithm, where the exact log-marginal likelihood and its gradient are calculated using the full Cholesky decomposition of $K_{\yvect}$ or $B$. 

Alongside these approaches for optimizing kernel parameters without approximation, we also evaluate the performance of approximate GP methods. For this experiment, we chose to compare against approximations found in the software package GPstuff \cite{vanhatalo2013gpstuff}, namely the fully and partial independent training conditional approaches (FITC, PITC), and the sparse variational GP (VAR) \cite{Titsias09}.
In order to match the computational cost of CG/PCG, which is in $\bigO(n^2)$, we set the number of inducing points for the approximate methods to be $n^{2/3}$. 
%% for a selection of the approximate methods considered in the previous section, but now as a means to obtain an approximate marginal likelihood. 
%% In particular, we report FITC, PITC, Subset of data, Block Jacobi, and Spectral GP. \noteMF{Report short acronyms that match what we will put in the figure}

All methods are initialized from the same set of kernel parameters, and % obtained by optimizing the log-marginal likelihood based on a subset of $4 \sqrt{n}$ points.
the curves are averaged over $5$ folds ($3$ for the Protein and EEG datasets).
For the sake of integrity, we ran each method in the comparison individually on a workstation with Intel Xeon E5-2630 CPU having 16 cores and 128GB RAM.
We also ensured that all methods reported in the comparison used optimized linear algebra routines exploiting the multi-core architecture.
This diligence for ensuring fairness gives credence to our assumption that the timings are not affected by external factors other than the actual implementation of the algorithms. The CG, PCG and CHOL approaches have been implemented in R; %, whereas the approximate methods were the ones implemented in GPstuff.
the fact that the approximate methods were implemented in a different environment (GPstuff is written in Matlab/Octave) and by a different developer may cast some doubt on the correctness of directly comparing results.
However, we believe that the key point emerging from this comparison is that preconditioning feasibly enables the use of iterative approaches for optimization of kernel parameters in GPs, and the results are competitive with those achieved using popular GP software packages.

For the reported experiments, it was possible to store the kernel matrix $K$ for all datasets, making it possible to compare methods against the baseline GP where computations use Cholesky decompositions.
We stress, however, that iterative approaches based on CG/PCG can be implemented without the need to store $K$, whereas this is not possible for approaches that attempt to factorize $K$ exactly. 
It is also worth noting that for the CG/PCG approach, calculating the log-likelihood on test data requires solving one linear system for each test point; this clearly penalizes the speed of these methods given the set-up of the experiment, where predictions are carried out every fixed number of iterations.

%% We make one final remark here on the fact that for smaller data sets the curves in fig.~\ref{fig:error_vs_time} for CG/PCG appear to be similar.
%% On closer inspection, we notice that this is due to the fact that most of the time is spent constructing $K$ and its derivatives with respect to kernel parameters.
%% PCG does allow for faster solution to the linear systems but this fact does not emerge clearly enough from the time analysis.

\section{Discussion and Conclusions}

Careful attention to numerical properties is essential in scaling machine learning to large and realistic datasets. 
Here we have introduced the use of preconditioning to the implementation of kernel machines, specifically, prediction and learning of kernel parameters for GPs.
Our novel scheme permits the use of any likelihood that factorizes over the data points, allowing us to tackle both regression and classification. 
We have shown robust performance improvements, in both accuracy and computational cost, over a host of state-of-the-art approximation methods for kernel machines. 
Notably, our method is exact in the limit of iterations, unlike approximate alternatives. 
We have also shown that the use of PCG is competitive with exact Cholesky decomposition in modestly sized datasets, when the Cholesky factors can be feasibly computed. 
When data and thus the kernel matrix grow large enough, Cholesky factorization becomes unfeasible, leaving PCG as the optimal choice.

One of the key features of a PCG implementation is that it does not require storage of any $\mathcal{O}(n^2)$ objects. 
We plan to extend our implementation to compute the elements of $K$ on the fly in one case, and in another case store $K$ in a distributed fashion (e.g. in TensorFlow/Spark). 
Furthermore, while we have focused on solving linear systems, we can also use preconditioning for other iterative algorithms involving the $K$ matrix, e.g., those to solve $\log(K) \vvect$ and $K^{1/2} \vvect$ \cite{Chen11}, as is often useful in estimating marginal likelihoods for probabilistic kernel models like GPs.

\newpage

\section*{Acknowledgements} 

KC and MF are grateful to Pietro Michiardi and Daniele Venzano for assisting the completion of this work by providing additional computational resources for running the experiments.
JPC acknowledges support from the Sloan Foundation, The Simons Foundation (SCGB\#325171 and SCGB\#325233), and The Grossman Center at Columbia University.
\bibliographystyle{icml2016_no_url}

%% \bibliography{filippone,additional_refs}

\begin{thebibliography}{32}
\providecommand{\natexlab}[1]{#1}
\providecommand{\url}[1]{\texttt{#1}}
\expandafter\ifx\csname urlstyle\endcsname\relax
  \providecommand{\doi}[1]{doi: #1}\else
  \providecommand{\doi}{doi: \begingroup \urlstyle{rm}\Url}\fi

\bibitem[Anitescu et~al.(2012)Anitescu, Chen, and Wang]{Anitescu12}
Anitescu, M., Chen, J., and Wang, L.
\newblock {A Matrix-free Approach for Solving the Parametric Gaussian Process
  Maximum Likelihood Problem}.
\newblock \emph{SIAM Journal on Scientific Computing}, 34\penalty0
  (1):\penalty0 A240--A262, 2012.

\bibitem[Ashby \& Falgout(1996)Ashby and Falgout]{Ashby96}
Ashby, S.~F. and Falgout, R.~D.
\newblock A {P}arallel {M}ultigrid {P}reconditioned {C}onjugate {G}radient
  algorithm for {G}roundwater {F}low {S}imulations.
\newblock \emph{Nuclear Science and Engineering}, 124\penalty0 (1):\penalty0
  145--159, 1996.

\bibitem[Asuncion \& Newman(2007)Asuncion and Newman]{Asuncion07}
Asuncion, A. and Newman, D.~J.
\newblock {UCI} machine learning repository, 2007.
\newblock URL \url{http://archive.ics.uci.edu/ml}.

\bibitem[Candela \& Rasmussen(2005)Candela and Rasmussen]{Candela05}
Candela, J.~Q. and Rasmussen, C.~E.
\newblock {A Unifying View of Sparse Approximate Gaussian Process Regression}.
\newblock \emph{Journal of Machine Learning Research}, 6:\penalty0 1939--1959,
  2005.

\bibitem[Chalupka et~al.(2013)Chalupka, Williams, and Murray]{Chalupka13}
Chalupka, K., Williams, C. K.~I., and Murray, I.
\newblock A framework for evaluating approximation methods for {G}aussian
  process regression.
\newblock \emph{Journal of Machine Learning Research}, 14\penalty0
  (1):\penalty0 333--350, 2013.

\bibitem[Chen et~al.(2011)Chen, Anitescu, and Saad]{Chen11}
Chen, J., Anitescu, M., and Saad, Y.
\newblock {Computing f(A)b via Least Squares Polynomial Approximations}.
\newblock \emph{SIAM Journal on Scientific Computing}, 33\penalty0
  (1):\penalty0 195--222, 2011.

\bibitem[Chen(2005)]{Chen05}
Chen, K.
\newblock \emph{Matrix Preconditioning Techniques and Applications}.
\newblock Cambridge Monographs on Applied and Computational Mathematics.
  Cambridge University Press, 2005.

\bibitem[Davies(2014)]{Davies14}
Davies, A.
\newblock \emph{{Effective Implementation of Gaussian Process Regression for
  Machine Learning}}.
\newblock PhD thesis, University of Cambridge, 2014.

\bibitem[Duchi et~al.(2011)Duchi, Hazan, and Singer]{Duchi11}
Duchi, J., Hazan, E., and Singer, Y.
\newblock {Adaptive Subgradient Methods for Online Learning and Stochastic
  Optimization}.
\newblock \emph{Journal of Machine Learning Research}, 12:\penalty0 2121--2159,
  2011.

\bibitem[Filippone \& Engler(2015)Filippone and Engler]{FilipponeICML15}
Filippone, M. and Engler, R.
\newblock Enabling scalable stochastic gradient-based inference for {G}aussian
  processes by employing the {U}nbiased {LI}near {S}ystem {S}olv{E}r
  ({ULISSE}).
\newblock In Blei, D. and Bach, F. (eds.), \emph{Proceedings of The 32nd
  International Conference on Machine Learning}, volume~37 of \emph{{JMLR}
  Proceedings}, pp.\  1015--1024, 2015.

\bibitem[Filippone et~al.(2013)Filippone, Zhong, and Girolami]{FilipponeML13}
Filippone, M., Zhong, M., and Girolami, M.
\newblock A comparative evaluation of stochastic-based inference methods for
  {G}aussian process models.
\newblock \emph{Machine Learning}, 93\penalty0 (1):\penalty0 93--114, 2013.

\bibitem[Flaxman et~al.(2015)Flaxman, Wilson, Neill, Nickisch, and
  Smola]{Flaxman15}
Flaxman, S., Wilson, A., Neill, D., Nickisch, H., and Smola, A.
\newblock Fast {K}ronecker inference in {G}aussian processes with
  non-{G}aussian likelihoods.
\newblock In Blei, D. and Bach, F. (eds.), \emph{Proceedings of The 32nd
  International Conference on Machine Learning}, volume~37 of \emph{{JMLR}
  Proceedings}, pp.\  607--616, 2015.

\bibitem[Gibbs(1997)]{GibbsPhD97}
Gibbs, M.~N.
\newblock \emph{{Bayesian} {G}aussian processes for regression and
  classification}.
\newblock PhD thesis, University of Cambridge, 1997.

\bibitem[Gilboa et~al.(2015)Gilboa, Saatci, and Cunningham]{Gilboa15}
Gilboa, E., Saatci, Y., and Cunningham, J.~P.
\newblock {Scaling Multidimensional Inference for Structured Gaussian
  Processes}.
\newblock \emph{{IEEE} Transactions on Pattern Analysis and Machine
  Intelligence}, 37\penalty0 (2):\penalty0 424--436, 2015.

\bibitem[Golub \& Van~Loan(1996)Golub and Van~Loan]{Golub96}
Golub, G.~H. and Van~Loan, C.~F.
\newblock \emph{{Matrix computations}}.
\newblock The Johns Hopkins University Press, 3rd edition, 1996.

\bibitem[Halko et~al.(2011)Halko, Martinsson, and Tropp]{Halko11}
Halko, N., Martinsson, P.~G., and Tropp, J.~A.
\newblock {Finding Structure with Randomness: Probabilistic Algorithms for
  Constructing Approximate Matrix Decompositions}.
\newblock \emph{SIAM Review}, 53\penalty0 (2):\penalty0 217--288, 2011.

\bibitem[Kuss \& Rasmussen(2005)Kuss and Rasmussen]{Kuss05}
Kuss, M. and Rasmussen, C.~E.
\newblock {Assessing Approximate Inference for Binary Gaussian Process
  Classification}.
\newblock \emph{Journal of Machine Learning Research}, 6:\penalty0 1679--1704,
  2005.

\bibitem[L\'{a}zaro-Gredilla et~al.(2010)L\'{a}zaro-Gredilla,
  Quinonero-Candela, Rasmussen, and Figueiras-Vidal]{Gredilla10}
L\'{a}zaro-Gredilla, M., Quinonero-Candela, J., Rasmussen, C.~E., and
  Figueiras-Vidal, A.~R.
\newblock {Sparse Spectrum Gaussian Process Regression}.
\newblock \emph{Journal of Machine Learning Research}, 11:\penalty0 1865--1881,
  2010.

\bibitem[Murray et~al.(2010)Murray, Adams, and MacKay]{Murray10b}
Murray, I., Adams, R.~P., and MacKay, D. J.~C.
\newblock Elliptical slice sampling.
\newblock In Teh, Y.~W. and Titterington, D.~M. (eds.), \emph{Proceedings of
  the Thirteenth International Conference on Artificial Intelligence and
  Statistics}, volume~9 of \emph{{JMLR} Proceedings}, pp.\  541--548, 2010.

\bibitem[Nickisch \& Rasmussen(2008)Nickisch and Rasmussen]{Nickisch08}
Nickisch, H. and Rasmussen, C.~E.
\newblock {Approximations for Binary Gaussian Process Classification}.
\newblock \emph{Journal of Machine Learning Research}, 9:\penalty0 2035--2078,
  2008.

\bibitem[Notay(2000)]{Notay00}
Notay, Y.
\newblock {Flexible Conjugate Gradients}.
\newblock \emph{SIAM Journal on Scientific Computing}, 22\penalty0
  (4):\penalty0 1444--1460, 2000.

\bibitem[Rahimi \& Recht(2008)Rahimi and Recht]{Rahimi08}
Rahimi, A. and Recht, B.
\newblock Random features for large-scale kernel machines.
\newblock In Platt, J.~C., Koller, D., Singer, Y., and Roweis, S.~T. (eds.),
  \emph{Advances in Neural Information Processing Systems 20}, pp.\
  1177--1184, 2008.

\bibitem[Rasmussen \& Williams(2006)Rasmussen and Williams]{Rasmussen06}
Rasmussen, C.~E. and Williams, C.
\newblock \emph{{Gaussian Processes for Machine Learning}}.
\newblock MIT Press, 2006.

\bibitem[Robbins \& Monro(1951)Robbins and Monro]{Robbins51}
Robbins, H. and Monro, S.
\newblock {A Stochastic Approximation Method}.
\newblock \emph{The Annals of Mathematical Statistics}, 22:\penalty0 400--407,
  1951.

\bibitem[Sch\"{o}lkopf \& Smola(2001)Sch\"{o}lkopf and Smola]{Scholkopf01}
Sch\"{o}lkopf, B. and Smola, A.~J.
\newblock \emph{{Learning with Kernels: Support Vector Machines,
  Regularization, Optimization, and Beyond}}.
\newblock MIT Press, Cambridge, MA, USA, 2001.

\bibitem[Snelson \& Ghahramani(2007)Snelson and Ghahramani]{Snelson07}
Snelson, E. and Ghahramani, Z.
\newblock {Local and global sparse Gaussian process approximations}.
\newblock In Meila, M. and Shen, X. (eds.), \emph{Proceedings of the 11th
  International Conference on Artificial Intelligence and Statistics}, volume~2
  of \emph{{JMLR} Proceedings}, pp.\  524--531, 2007.

\bibitem[Srinivasan et~al.(2014)Srinivasan, Hu, Gumerov, Murtugudde, and
  Duraiswami]{Srinivasan14}
Srinivasan, B.~V., Hu, Q., Gumerov, N.~A., Murtugudde, R., and Duraiswami, R.
\newblock {Preconditioned Krylov solvers for kernel regression}, 2014.
\newblock arXiv:1408.1237.

\bibitem[Stein et~al.(2012)Stein, Chen, and Anitescu]{Stein12}
Stein, M.~L., Chen, J., and Anitescu, M.
\newblock Difference {F}ilter {P}reconditioning for {L}arge {C}ovariance
  {M}atrices.
\newblock \emph{SIAM Journal on Matrix Analysis Applications}, 33\penalty0
  (1):\penalty0 52--72, 2012.

\bibitem[Titsias(2009)]{Titsias09}
Titsias, M.~K.
\newblock {Variational Learning of Inducing Variables in Sparse Gaussian
  Processes}.
\newblock In Dyk, D.~A. and Welling, M. (eds.), \emph{Proceedings of the 12th
  International Conference on Artificial Intelligence and Statistics}, volume~5
  of \emph{{JMLR} Proceedings}, pp.\  567--574, 2009.

\bibitem[Vanhatalo et~al.(2013)Vanhatalo, Riihim{\"a}ki, Hartikainen,
  Jyl{\"a}nki, Tolvanen, and Vehtari]{vanhatalo2013gpstuff}
Vanhatalo, J., Riihim{\"a}ki, J., Hartikainen, J., Jyl{\"a}nki, P., Tolvanen,
  V., and Vehtari, A.
\newblock {GP}stuff: {B}ayesian modeling with {G}aussian processes.
\newblock \emph{Journal of Machine Learning Research}, 14\penalty0
  (1):\penalty0 1175--1179, 2013.

\bibitem[Williams \& Seeger(2000)Williams and Seeger]{Williams00b}
Williams, C. K.~I. and Seeger, M.~W.
\newblock Using the {N}ystr{\"{o}}m method to speed up kernel machines.
\newblock In Leen, T.~K., Dietterich, T.~G., and Tresp, V. (eds.),
  \emph{Advances in Neural Information Processing Systems 13}, pp.\  682--688,
  2000.

\bibitem[Wilson \& Nickisch(2015)Wilson and Nickisch]{Wilson15}
Wilson, A. and Nickisch, H.
\newblock {Kernel Interpolation for Scalable Structured Gaussian Processes
  (KISS-GP)}.
\newblock In Blei, D. and Bach, F. (eds.), \emph{Proceedings of the 32nd
  International Conference on Machine Learning}, volume~37 of \emph{{JMLR}
  Proceedings}, pp.\  1775--1784, 2015.

\end{thebibliography}

%% \twocolumn[{%
%% \centering
%% \vspace{0.5cm}
%% {\Large \bf Supplementary material}
%% \vspace{1cm}
%% }]

%% \end{document} 

\appendix

\twocolumn[{}]

\section{Other results not included in the paper}

\begin{figure}[h!]
% \vskip 0.2in
\begin{center}
\begin{tabular}{cc}
%% \multicolumn{4}{c}{{\bf Concrete dataset - regression - $n \approx 1K$}} \\
%% \multicolumn{2}{c}{{\bf Isotropic}} &
%% \multicolumn{2}{c}{{\bf ARD}} \\
\hspace{-0.4cm} \includegraphics[width=\columnwidth/2]{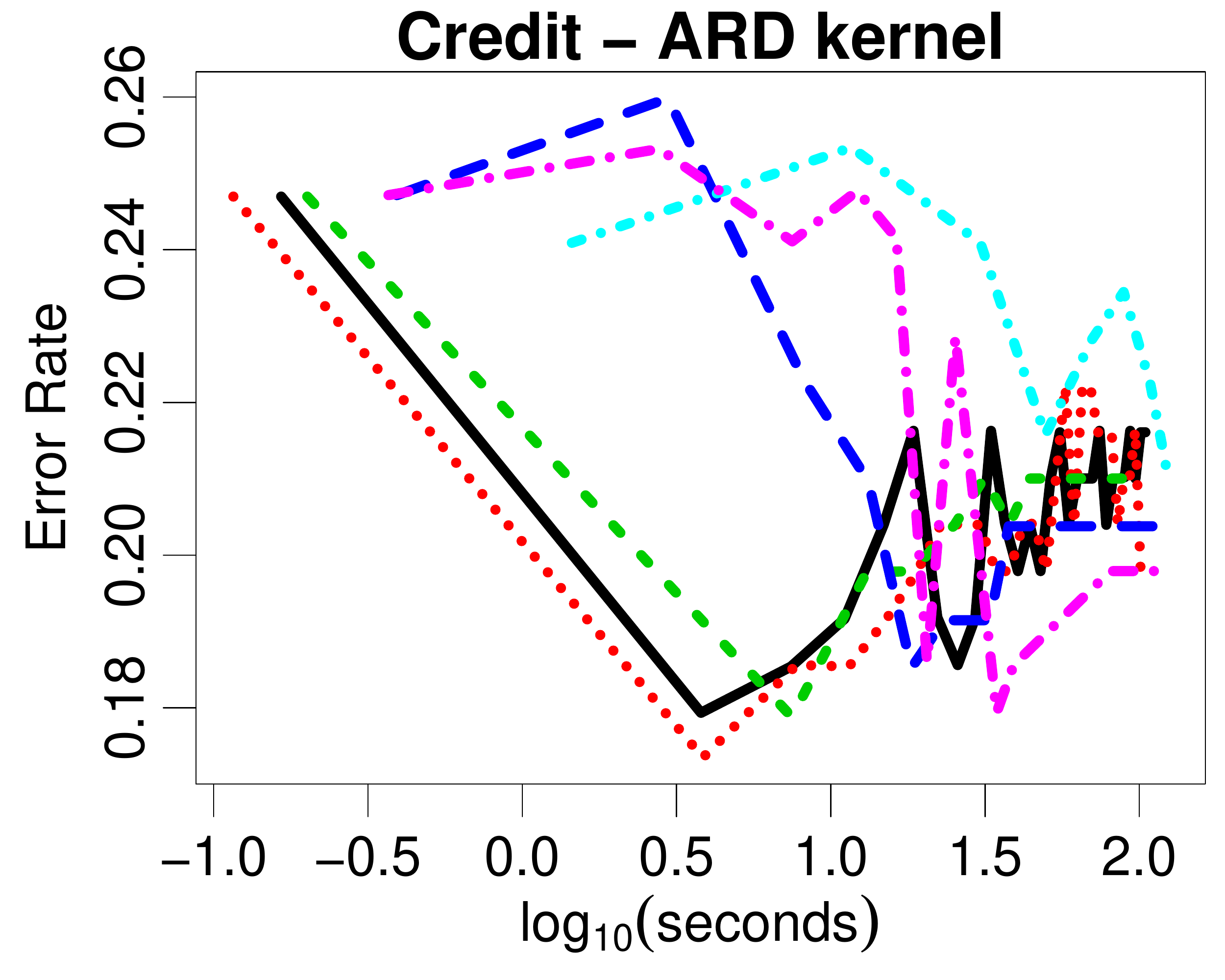} \hspace{-0.7cm} & \includegraphics[width=\columnwidth/2]{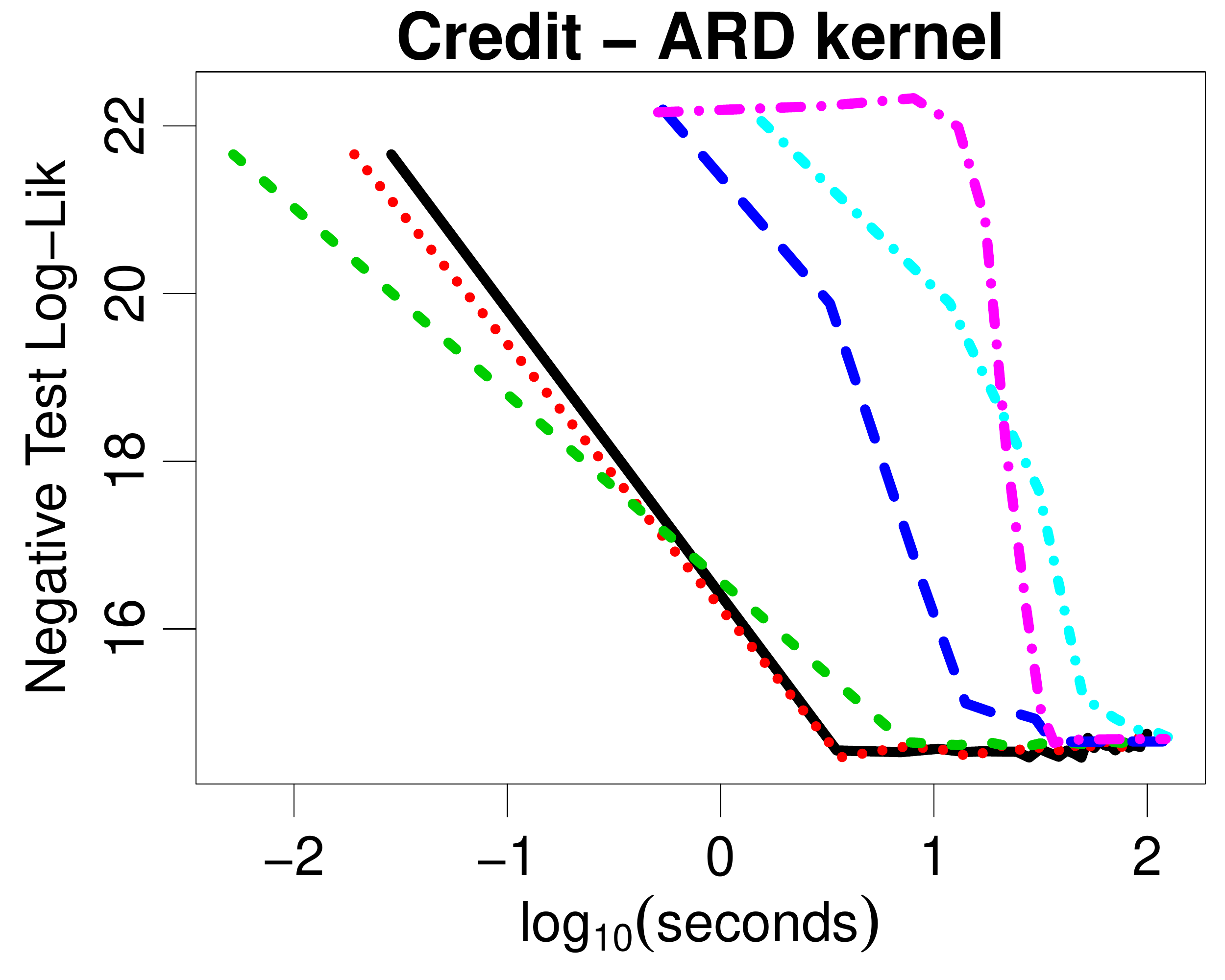} \\
\hspace{-0.4cm} \includegraphics[width=\columnwidth/2]{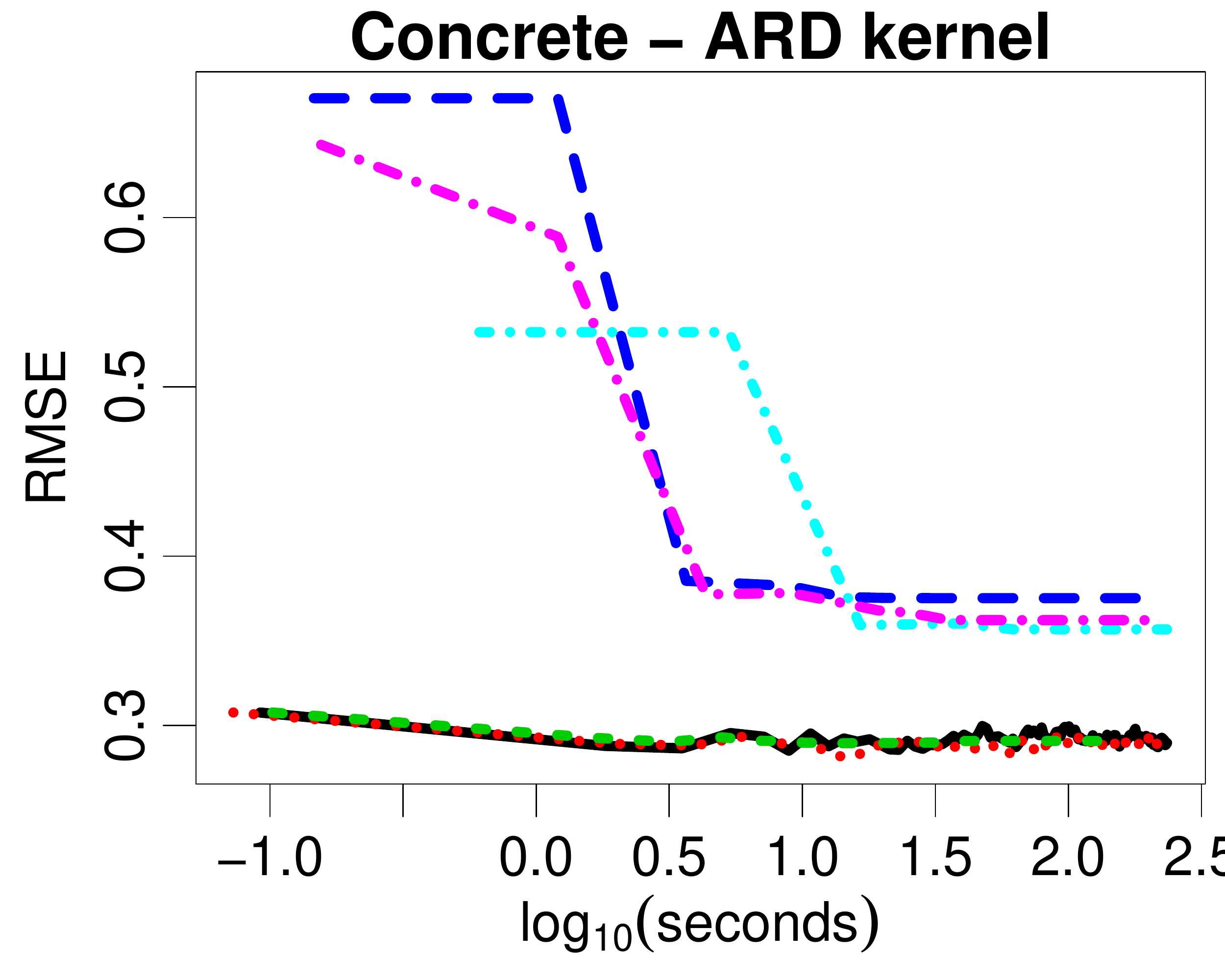} \hspace{-0.7cm} & \includegraphics[width=\columnwidth/2]{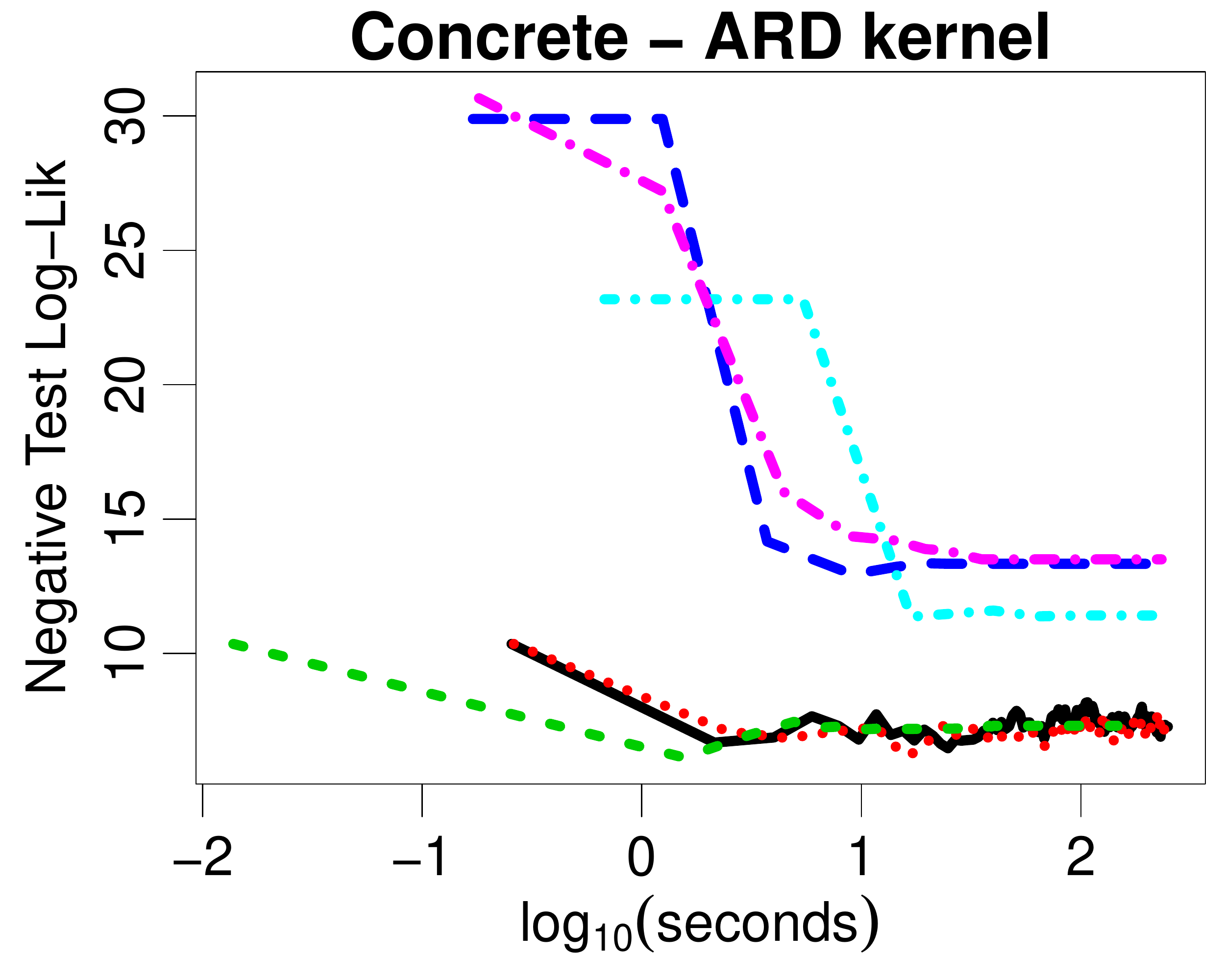} \\
\multicolumn{2}{c}{\includegraphics[width=\columnwidth]{Plots/PLOT_COMPARE_ERROR_VS_TIME_LEGEND.pdf}} 
\end{tabular}
\caption{RMSE and negative log of the likelihood on $\sqrt{n}$ held out test data over time. GP models employ the ARD kernel in eq.~\ref{eq:rbf:covariance}. GP classification: Credi dataset ($n=1000, d=24$). GP regression: Concrete dataset ($n=1029, d=8$). Curves are averaged over multiple repetitions.}
\label{fig:error_vs_time:supplement}
\end{center}
\vskip -0.2in
\end{figure}

In fig.~\ref{fig:error_vs_time:supplement} we report some of the runs that we did not include in the main text for lack of space.
The figure reports plots on the error vs.\ time for the same regression cases considered in the main text but with an isotropic kernel, and results on the concrete dataset with isotropic and ARD kernels.

\section{Gaussian Processes with non-Gaussian likelihood functions}

In this section we report the derivations of the quantities needed to compute an unbiased estimate of the log-marginal likelihood given by the Laplace approximation for GP models with non-Gaussian likelihood functions.
Throughout this section, we assume a factorizing likelihood 
$$
p(\yvect | \fvect) = \prod_{i=1}^n p(y_i | f_i).
$$
and we specialize the equations to the probit likelihood
\begin{equation}
p(y_i \mid f_i) = % \Phi(f_i)^{y_i} (1 - \Phi(f_i))^{(1 - y_i)/2} = 
\Phi(y_i f_i).
\end{equation}
where $\Phi$ denotes the cumulative function of the Gaussian density.
The latent variables $\fvect$ are given a zero mean GP prior $\fvect \sim \norm(\fvect | \zerovect, K)$.

For a given value of the hyperparameters $\thetavect$, define
\begin{equation} \label{eq:laplace:target}
\Psi(\fvect) = \log[p(\yvect \mid \fvect)] + \log[p(\fvect \mid \thetavect)] + \mathrm{const.}
\end{equation}
as the logarithm of the posterior density over $\fvect$.
Performing a Laplace approximation amounts in defining a Gaussian $q(\fvect \mid \yvect, \thetavect) = \norm(\fvect \mid \hat{\fvect}, \hat{\Sigma})$, such that
% $$
% p(\fvect \mid \yvect, \thetavect) \simeq q(\fvect \mid \thetavect) = \norm(\hat{\fvect}, \Sigma_{\fvect})
% $$
% where
\begin{equation} \label{eq:fhat}
\hat{\fvect} = \arg\max_{\fvect} \Psi(\fvect) \qquad \mathrm{and} \qquad
\hat{\Sigma}^{-1} = - \nabla_{\fvect} \nabla_{\fvect} \Psi(\hat{\fvect}).
\end{equation}
As it is not possible to directly solve the maximization problem in equation~\ref{eq:fhat}, an iterative procedure based on the following Newton-Raphson formula is usually employed,
\begin{equation} \label{eq:laplace:newton:iteration}
% \fvect^{(k+1)} = \fvect^{(k)} - (\nabla_{\fvect} \nabla_{\fvect} \Psi(\fvect^{(k)}))^{-1} \nabla_{\fvect} \Psi(\fvect^{(k)})
\fvect^{\mathrm{new}} = \fvect - (\nabla_{\fvect} \nabla_{\fvect} \Psi(\fvect))^{-1} \nabla_{\fvect} \Psi(\fvect),
\end{equation}
starting from some initial $\fvect$ until convergence.
The gradient and the Hessian of the log of the target density are
\begin{equation}
\nabla_{\fvect} \Psi(\fvect) = \nabla_{\fvect} \log[p(\yvect \mid \fvect)] - K^{-1} \fvect \text{\quad and }
\end{equation}
\begin{equation}
\nabla_{\fvect} \nabla_{\fvect} \Psi(\fvect) = \nabla_{\fvect} \nabla_{\fvect} \log[p(\yvect \mid \fvect)] - K^{-1} = -W - K^{-1},
\end{equation}
where we have defined $W = - \nabla_{\fvect} \nabla_{\fvect} \log[p(\yvect \mid \fvect)]$,
which is diagonal because the likelihood factorizes over observations.
Note that if $\log[p(\yvect \mid \fvect)]$ is concave, such as in probit classification, $\Psi(\fvect)$ has a unique maximum.

Standard manipulations lead to
$$
\fvect^{\mathrm{new}} = (K^{-1} + W)^{-1} (W \fvect + \nabla_{\fvect} \log[p(\yvect \mid \fvect)]).
$$

We can rewrite the inverse of the negative Hessian using the matrix inversion lemma:
$$
\left(K^{-1} + W\right)^{-1} = K - K W^{\frac{1}{2}} B^{-1} W^{\frac{1}{2}} K,
$$
where
$$
B = I + W^{\frac{1}{2}} K W^{\frac{1}{2}}.
$$
This means that each iteration becomes:
$$
\fvect^{\mathrm{new}} = (K - K W^{\frac{1}{2}} B^{-1} W^{\frac{1}{2}} K) (W \fvect + \nabla_{\fvect} \log[p(\yvect \mid \fvect)]).
$$ 
We can define $\bvect = (W \fvect + \nabla_{\fvect} \log[p(\yvect \mid \fvect)])$ and rewrite this expression as:
%% $$
%% \fvect^{\mathrm{new}} = K (I - W^{\frac{1}{2}} B^{-1} W^{\frac{1}{2}} K) \bvect.
%% $$
%% Finally, we can rewrite
$$
\fvect^{\mathrm{new}} = K (\bvect - W^{\frac{1}{2}} B^{-1} W^{\frac{1}{2}} K \bvect).
$$
From this, we see that at convergence
$$
\avect = K^{-1} \hat{\fvect} = (\bvect - W^{\frac{1}{2}} B^{-1} W^{\frac{1}{2}} K \bvect).
$$
As we will see later, the definition of $\avect$ is useful for the calculation of the gradient and for predictions.

Proceeding with the calculations from right to left we see that in order to complete a Newton-Raphson iteration the expensive operations are:
(i) carry out one matrix-vector multiplication $K \bvect$,
(ii) solve a linear system involving the $B$ matrix,
and (iii) carry out one matrix-vector multiplication involving $K$ and the vector in the parenthesis.
Calculating $\bvect$ and performing any multiplications of $W^{\frac{1}{2}}$ with vectors cost $\bigO(n)$.

All these operations can be carried out without the need to store $K$ or any other $n \times n$ matrices.
The linear system in (ii) can be solved using the CG algorithm that involves repeatedly multiplying $B$ (and therefore $K$) with vectors.

%% It is interesting to notice that the matrix $B$ is well behaved, as its smallest eigenvalue is grater than one, and its largest eigenvalue is smaller than $1 + n \max(k_{ij})/4$

\begin{algorithm}[tb]
   \caption{Laplace approximation for GPs}
   \label{alg:stochastic:gradient:laplace}
\begin{algorithmic}[1]
   \STATE {\bfseries Input:} data $X$, labels $\yvect$, likelihood function $p(\yvect \mid \fvect)$
   \STATE $\fvect = \zerovect$ 
   \REPEAT
   \STATE Compute $\diag(W)$, $\bvect$, $W^{\frac{1}{2}} K \bvect$
   \STATE $\mathrm{solve}(B, W^{\frac{1}{2}} K \bvect)$
   \STATE Compute $\avect$, $K \avect$
   \STATE Compute $\fvect^{\mathrm{new}}$
   \UNTIL{ convergence}
   \STATE {\bf return} $\hat{\fvect}$, $\avect$ 
\end{algorithmic}
\end{algorithm}

\subsection{Stochastic gradients}

The Laplace approximation yields an approximate log-marginal likelihood in the following form:
\begin{equation}
\log[\hat{p}(\yvect \mid \thetavect, X)] = -\frac{1}{2} \log|B| - \frac{1}{2} \hat{\fvect}^{\T} K^{-1} \hat{\fvect} + \log[p(\yvect \mid \hat{\fvect})]
\end{equation}
Handy relationships that we will be using in the remainder of this section are:
$$
\log|B| = \log|I + W^{\frac{1}{2}}KW^{\frac{1}{2}}| = \log|I + KW|;
$$
$$
(I + KW)^{-1} = W^{-\frac{1}{2}} B^{-1} W^{\frac{1}{2}}.
$$

The gradient of the log-marginal likelihood with respect to the kernel parameters $\thetavect$ requires differentiating the terms that explicitly depend on $\thetavect$ and those that implicitly depend on it because a change in the parameters reflects in a change in $\hat{\fvect}$.
Denoting by $g_i$ the $i$th component of the gradient of $\frac{\partial \log[\hat{p}(\yvect | \thetavect)]}{\partial \theta_i}$, we obtain
\begin{eqnarray}
g_i & = & % \frac{\partial \log[\hat{p}(\yvect | \thetavect)]}{\partial \theta_i} & = & 
- \frac{1}{2} \Tr \left( B^{-1} \frac{\partial B}{\partial \theta_i} \right) \nonumber \\
& & + \frac{1}{2} \hat{\fvect}^{\T} K^{-1} \frac{\partial K}{\partial \theta_i} K^{-1} \hat{\fvect} \nonumber  \\
& & + \left[\nabla_{\hat{\fvect}}  \log[\hat{p}(\yvect | \thetavect)] \right]^{\T} \frac{\partial \hat{\fvect}}{\partial \theta_i} 
\end{eqnarray}

The trace term cannot be computed exactly for large $n$ so we propose a stochastic estimate:
$$
- \frac{1}{2} \widetilde{\left[\Tr \left( B^{-1} \frac{\partial B}{\partial \theta_i} \right) \right]} = 
- \frac{1}{2 N_{\rvect}} \sum_{i=1}^{N_{\rvect}}  (\rvect^{(i)})^{\T} B^{-1} \frac{\partial B}{\partial \theta_i} \rvect^{(i)}.
$$ 
By noticing that the derivative of $B$ is $W^{\frac{1}{2}}\frac{\partial K}{\partial \theta_i}W^{\frac{1}{2}}$, this simplifies to
$$
%% - \frac{1}{2} \widetilde{\left[ \Tr \left( B^{-1} \frac{\partial B}{\partial \theta_i} \right) \right]} = 
- \frac{1}{2 N_{\rvect}} \sum_{i=1}^{N_{\rvect}}  (\rvect^{(i)})^{\T} B^{-1} W^{\frac{1}{2}} \frac{\partial K}{\partial \theta_i} W^{\frac{1}{2}} \rvect^{(i)},
$$ 
so we need to solve $N_{\rvect}$ linear systems involving $B$.

The second term contains the linear system $K^{-1} \hat{\fvect}$ that we already have from the Laplace approximation and is $\avect$.

The third term is slightly more involved and will be dealt with in the next sub-section.

\begin{algorithm}[tb]
   \caption{Stochastic gradients for GPs}
   \label{alg:stochastic:gradient:laplace}
\begin{algorithmic}[1]
   \STATE {\bfseries Input:} data $X$, labels $\yvect$, $\hat{\fvect}$, $\avect$
   \STATE $\mathrm{solve}(B, \rvect^{(i)})$ for $i=1, \ldots, N_{\rvect}$
   \STATE Compute first term of $\tilde{g}_i$
   \STATE Compute second term of $\tilde{g}_i$
   \STATE $\mathrm{solve}(B, W^{\frac{1}{2}} K \rvect^{(i)})$ for $i=1, \ldots, N_{\rvect}$
   \STATE Compute $\tilde{\uvect}$
   \STATE $\mathrm{solve}(B, W^{\frac{1}{2}} \frac{\partial K}{\partial \theta_i} \nabla_{\hat{\fvect}} \log[p(\yvect \mid \hat{\fvect})])$
   \STATE Compute third term of $\tilde{g}_i$
   \STATE {\bf return} $\tilde{\gvect}$ 
\end{algorithmic}
\end{algorithm}

\subsubsection{Implicit derivatives}
The last (implicit) term in the last equation can be simplified by noticing that:
$$
\log[\hat{p}(\yvect \mid \thetavect)] = \Psi(\hat{\fvect}) - \frac{1}{2} \log|B|
$$
and that the derivative of the first term wrt $\hat{\fvect}$ is zero because $\hat{\fvect}$ maximizes $\Psi(\hat{\fvect})$.
Therefore:
$$
\left[\nabla_{\hat{\fvect}}  \log[\hat{p}(\yvect \mid \thetavect)] \right]^{\T} \frac{\partial \hat{\fvect}}{\partial \theta_i}
=
- \frac{1}{2} \left[\nabla_{\hat{\fvect}}  \log|B| \right]^{\T} \frac{\partial \hat{\fvect}}{\partial \theta_i}
$$
The components of $\left[\nabla_{\hat{\fvect}}  \log|B| \right]$ can be obtained by considering the identity $\log|B| = \log|I + KW|$, so differentiating $\log|B|$ wrt the components of $\hat{\fvect}$ becomes:
$$
\frac{\partial \log|I + KW| }{\partial (\hat{\fvect})_j } = \Tr\left( (I + KW)^{-1} K \frac{\partial W}{\partial (\hat{\fvect})_j}  \right)
$$
We can rewrite this by gathering $K$ inside the inverse and, due to the inversion of the matrix product, $K$ cancels out:
$$
\frac{\partial \log|I + KW| }{\partial (\hat{\fvect})_j } = \Tr\left( (K^{-1} + W)^{-1} \frac{\partial W}{\partial (\hat{\fvect})_j}  \right)
$$
We notice here that the resulting trace contains the inverse of the same matrix needed in the iterations of the Laplace approximation and that the matrix $\frac{\partial W}{\partial (\hat{\fvect})_j}$ is zero everywhere except in the $j$th diagonal element where it attains the value:
$$
\frac{\partial W}{\partial (\hat{\fvect})_j} = \frac{\partial^3 \log[p(\yvect \mid \hat{\fvect})]}{\partial (\hat{\fvect})^3_j}
$$
For this reason, it would be possible to simplify the trace term as the product between the $j$th diagonal element of $(K^{-1} + W)^{-1}$ and $\frac{\partial^3 \log[p(\yvect \mid \hat{\fvect})]}{\partial (\hat{\fvect})^3_j}$.
Bearing in mind that we need $n$ of these quantities, we could define
$$
D = \diag\left[\diag\left[ (K^{-1} + W)^{-1} \right]\right]
$$
$$
(\dvect)_j = \frac{\partial^3 \log[p(\yvect \mid \hat{\fvect})]}{\partial (\hat{\fvect})^3_j}
$$
and rewrite
$$
- \frac{1}{2} \left[\nabla_{\hat{\fvect}}  \log|B| \right] = - \frac{1}{2} D \dvect
$$
which is the standard way to proceed when computing the gradient of the approximate log-marginal likelihood using the Laplace approximation \cite{Rasmussen06}.
However, this would be difficult to compute exactly for large $n$, as this would require inverting $K^{-1} + W$ first and then compute its diagonal.
Using the matrix inversion lemma would not simplify things as there would still be an inverse of $B$ to compute explicitly.
We therefore aim for a stochastic estimate of this term starting from:
\begin{eqnarray}
\frac{\partial \log|I + KW| }{\partial (\hat{\fvect})_j } & = & \Tr\left( (K^{-1} + W)^{-1} \frac{\partial W}{\partial (\hat{\fvect})_j}  \right) \nonumber \\
& = & \Tr\left( (K^{-1} + W)^{-1} \frac{\partial W}{\partial (\hat{\fvect})_j} \E[\rvect \rvect^{\T}]  \right) \nonumber \\
\end{eqnarray}
where we have introduced the $\rvect$ vectors with the property $\E[\rvect \rvect^{\T}] = I$.
So an unbiased estimate of the trace for each component of $\hat{\fvect}$ is:
\begin{eqnarray}
(\tilde{\uvect})_j & = & \widetilde{\left[\frac{\partial \log|I + KW| }{\partial (\hat{\fvect})_j }\right]} \nonumber \\
& = & \frac{1}{N_{\rvect}} \sum_{i=1}^{N_{\rvect}}  (\rvect^{(i)})^{\T} (K^{-1} + W)^{-1} \frac{\partial W}{\partial (\hat{\fvect})_j} \rvect^{(i)}   \nonumber \\
\end{eqnarray}
which requires solving $N_{\rvect}$ linear systems involving the $B$ matrix:
$$
(K^{-1} + W)^{-1} \rvect^{(i)} =
K (\rvect^{(i)} - W^{\frac{1}{2}} B^{-1} W^{\frac{1}{2}} K \rvect^{(i)})
$$

The derivative of $\hat{\fvect}$ wrt $\theta_i$ can be obtained by differentiating the expression $\hat{\fvect} = K \nabla_{\hat{\fvect}} \log[p(\yvect \mid \hat{\fvect})]$:
$$
\frac{\partial \hat{\fvect}}{\partial \theta_i} = \frac{\partial K}{\partial \theta_i} \nabla_{\hat{\fvect}} \log[p(\yvect \mid \hat{\fvect})] + K \nabla_{\hat{\fvect}} \nabla_{\hat{\fvect}} \log[p(\yvect \mid \hat{\fvect})] \frac{\partial \hat{\fvect}}{\partial \theta_i}
$$
Given that $\nabla_{\hat{\fvect}} \nabla_{\hat{\fvect}} \log[p(\yvect \mid \hat{\fvect})] = -W$ we can rewrite:
$$
 (I + KW) \frac{\partial \hat{\fvect}}{\partial \theta_i} = \frac{\partial K}{\partial \theta_i} \nabla_{\hat{\fvect}} \log[p(\yvect \mid \hat{\fvect})]
$$
which yields:
$$
\frac{\partial \hat{\fvect}}{\partial \theta_i} = (I + KW)^{-1} \frac{\partial K}{\partial \theta_i} \nabla_{\hat{\fvect}} \log[p(\yvect \mid \hat{\fvect})]
$$

So an unbiased estimate of the implicit term in the gradient of the approximate log-marginal likelihood becomes:
$$
- \frac{1}{2} \tilde{\uvect}^{\T} (I + KW)^{-1} \frac{\partial K}{\partial \theta_i} \nabla_{\hat{\fvect}} \log[p(\yvect \mid \hat{\fvect})]
$$
Rewriting the inverse in terms of $B$ yields:
$$
- \frac{1}{2} \tilde{\uvect}^{\T} W^{-\frac{1}{2}} B^{-1} W^{\frac{1}{2}} \frac{\partial K}{\partial \theta_i} \nabla_{\hat{\fvect}} \log[p(\yvect \mid \hat{\fvect})]
$$

Putting everything together, the components of the stochastic gradient are:
\begin{eqnarray}
%% \widetilde{\left[\frac{\partial \log[\hat{p}(\yvect \mid \thetavect)]}{\partial \theta_i} \right]} & = & 
\tilde{g}_i & = & 
- \frac{1}{2 N_{\rvect}} \sum_{i=1}^{N_{\rvect}}  (\rvect^{(i)})^{\T} B^{-1} W^{\frac{1}{2}} \frac{\partial K}{\partial \theta_i} W^{\frac{1}{2}} \rvect^{(i)} \nonumber \\
& & + \frac{1}{2} \avect^{\T} \frac{\partial K}{\partial \theta_i} \avect \nonumber  \\
& & - \frac{1}{2} \tilde{\uvect}^{\T} W^{-\frac{1}{2}} B^{-1} W^{\frac{1}{2}} \frac{\partial K}{\partial \theta_i} \nabla_{\hat{\fvect}} \log[p(\yvect | \hat{\fvect})]
\end{eqnarray}

\subsection{Predictions}
%%% WARNING - THIS IS SPECIFIC FOR CLASSIFICATION ONLY!! MAYBE WE SHOULD REWRITE THE LAST LINES OF THE ALGORITHM %%%%
\begin{algorithm}[tb]
   \caption{Prediction for GPs with Laplace approximation without Cholesky decompositions}
   \label{alg:stochastic:gradient:laplace}
\begin{algorithmic}[1]
   \STATE {\bfseries Input:} data $X$, labels $\yvect$, test input $\xvect_*$, $\hat{\fvect}$, $\avect$
   \STATE Compute $\mu_*$ 
   \STATE $\mathrm{solve}(B, W^{\frac{1}{2}} \kvect_*)$
   \STATE Compute $s^2_*$, $\Phi\left( \frac{m_*}{\sqrt{1 + s^2_*}} \right)$
   \STATE {\bf return} $\Phi\left( \frac{m_*}{\sqrt{1 + s^2_*}} \right)$ 
\end{algorithmic}
\end{algorithm}

To obtain an approximate predictive distribution, conditioned on a value of the hyperparameters $\thetavect$, we can compute:
\begin{equation} \label{eq:approx:integration:f}
p(y_* \mid \yvect, \thetavect) = \int p(y_* \mid f_*) p(f_* \mid \fvect, \thetavect) q(\fvect \mid \yvect, \thetavect) df_* d\fvect.
\end{equation}
%% Here $\thetavect$ can be a ML estimate that maximizes the approximate likelihood or one sample from the approximate posterior $\hat{p}(\thetavect | \yvect)$.

Given the properties of multivariate normal variables, $f_*$ is distributed as $\norm(f_* \mid \mu_*, \beta^2_*)$ with $\mu_* = \kvect_*^{\T} K^{-1} \fvect$ and $\beta^2_* = k_{**} - \kvect_*^{\T} K^{-1} \kvect_*$.
Approximating $p(\fvect \mid \yvect, \thetavect)$ with a Gaussian $q(\fvect \mid \yvect, \thetavect) = \norm(\fvect \mid \muvect_q, \Sigma_q)$ makes it possible to analytically perform integration with respect to $\fvect$ in eq. \ref{eq:approx:integration:f}.
In particular, the integration with respect to $\fvect$ yields $\norm(f_* \mid m_*, s^2_*)$ with
$$
m_* = \kvect_*^{\T} K^{-1} \hat{\fvect}
$$
and 
$$
s^2_* = k_{**} - \kvect_*^{\T} (K + W^{-1})^{-1} \kvect_*
$$
These quantities can be rewritten as:
$$
m_* = \kvect_*^{\T} \avect
$$
and 
$$
s^2_* = k_{**} - \kvect_*^{\T} W^{\frac{1}{2}} B^{-1} W^{\frac{1}{2}} \kvect_*
$$
This shows that the mean is cheap to compute, whereas the variance requires solving another linear system involving $B$ for each test point.

The univariate integration with respect to $f_*$ follows exactly in the case of a probit likelihood, as it is a convolution of a Gaussian and a cumulative Gaussian %~\cite{Rasmussen06} % using quadrature techniques or by means of the Monte Carlo estimate:
\begin{equation}
\int p(y_* \mid f_*) \norm(f_* \mid m_*, s^2_*) df_* = \Phi\left( \frac{m_*}{\sqrt{1 + s^2_*}} \right).
\end{equation}
% \begin{equation}
% \int p(y_* \mid f_*) p(f_* \mid \fvect, \thetavect) df_* = \Phi\left( \frac{m_*}{\sqrt{1 + s^2_*}} \right).
% \end{equation}

\subsection{Low rank preconditioning}

When a low rank approximation of the matrix $K$ is available, say $\hat{K} = \Phi \Phi^{\T}$, the inverse of the preconditioner can be rewritten as:
$$
(I + W^{\frac{1}{2}} \hat{K} W^{\frac{1}{2}})^{-1} = (I + W^{\frac{1}{2}} \Phi \Phi^{\T} W^{\frac{1}{2}})^{-1}
$$
By using the matrix inversion lemma we obtain:
$$
(I + W^{\frac{1}{2}} \Phi \Phi^{\T} W^{\frac{1}{2}})^{-1} = I - W^{\frac{1}{2}} \Phi (I + \Phi^{\T} W \Phi)^{-1} \Phi^{\T} W^{\frac{1}{2}}
$$
Similarly to the GP regression case, the application of this preconditioner is in $\bigO(m^3)$, where $m$ is the rank of $\Phi$.

\end{document}